\documentclass[runningheads]{llncs}

\usepackage{eccv}

\usepackage{eccvabbrv}

\usepackage{graphicx}
\usepackage{booktabs}

\usepackage[accsupp]{axessibility}  % Improves PDF readability for those with disabilities.

\usepackage{hyperref}

\usepackage{orcidlink}

\usepackage{array,multirow,graphicx}
\usepackage{float}
\usepackage{amsmath}
\usepackage{mathtools}
\usepackage{bbm}
\usepackage{hhline}
\usepackage{boldline}

\usepackage{kotex}
\usepackage{xcolor,colortbl}
\usepackage{nicematrix}
\usepackage{wrapfig}

\begin{document}

% \title{Trustworthy Contrastive Learning via Progressive Anchor Propagation for Unsupervised Semantic Segmentation} 
\title{Progressive Proxy Anchor Propagation for Unsupervised Semantic Segmentation}

\titlerunning{Progressive Proxy Anchor Propagation}

\author{Hyun Seok Seong\orcidlink{0000-0002-7952-2017} \and
WonJun Moon\orcidlink{0000-0003-2805-0926} \and
SuBeen Lee\orcidlink{0009-0005-1470-1160} \and
Jae-Pil Heo\thanks{Corresponding author}\orcidlink{0000-0001-9684-7641}
}

\authorrunning{H.S. Seong et al.}

\institute{Sungkyunkwan University \\
% \email{\{gustjrdl95,wjun0830,leesb7426,jaepilheo\}@skku.edu}
}

\maketitle

\begin{abstract}
The labor-intensive labeling for semantic segmentation has spurred the emergence of Unsupervised Semantic Segmentation.
Recent studies utilize patch-wise contrastive learning based on features from image-level self-supervised pretrained models.
However, relying solely on similarity-based supervision from image-level pretrained models often leads to unreliable guidance due to insufficient patch-level semantic representations.
To address this, we propose a Progressive Proxy Anchor Propagation~(PPAP) strategy.
This method gradually identifies more trustworthy positives for each anchor by relocating its proxy to regions densely populated with semantically similar samples.
Specifically, we initially establish a tight boundary to gather a few reliable positive samples around each anchor.
Then, considering the distribution of positive samples, we relocate the proxy anchor towards areas with a higher concentration of positives and adjust the positiveness boundary based on the propagation degree of the proxy anchor.
% In addition, there might exist ambiguous regions where positive and negative samples coexist near the positiveness boundary.
% Therefore, to further ensure the reliability of the negative set, we define an instance-wise ambiguous zone and exclude samples in such regions from the negative set.
Moreover, to account for ambiguous regions where positive and negative samples may coexist near the positiveness boundary, we introduce an instance-wise ambiguous zone. Samples within these zones are excluded from the negative set, further enhancing the reliability of the negative set.
Our state-of-the-art performances on various datasets validate the effectiveness of the proposed method
for Unsupervised Semantic Segmentation.
Our code is available at \href{https://github.com/hynnsk/PPAP}{https://github.com/hynnsk/PPAP}.

\keywords{Unsupervised \!Semantic \!Segmentation, \!Contrastive \!Learning}
\end{abstract}    
\section{Introduction}
\label{sec_introduction}

Semantic Segmentation plays a vital role in various fields, including robotics and autonomous driving~\cite{deeplab, dilated, segformer, segmenter, pyramidpool1, cnnseg2, refinenet, danet}.
With the abundant data available in media, developing high-quality semantic segmentation models has become feasible~\cite{sam}, though this has also increased the demand for extensive human annotations. 
Likewise, the increasing burden on human labor has spurred the emergence of Unsupervised Semantic Segmentation~(USS)~\cite{iic, picie, stego, transfgu, hp, cause, seitzer2023bridging, depthg, eagle, equss}.

\begin{figure}[t]
  \centering
  \begin{subfigure}{0.7\linewidth}
    % \fbox{\rule{0pt}{0.5in} \rule{.9\linewidth}{0pt}}
        \includegraphics[width=1\textwidth]{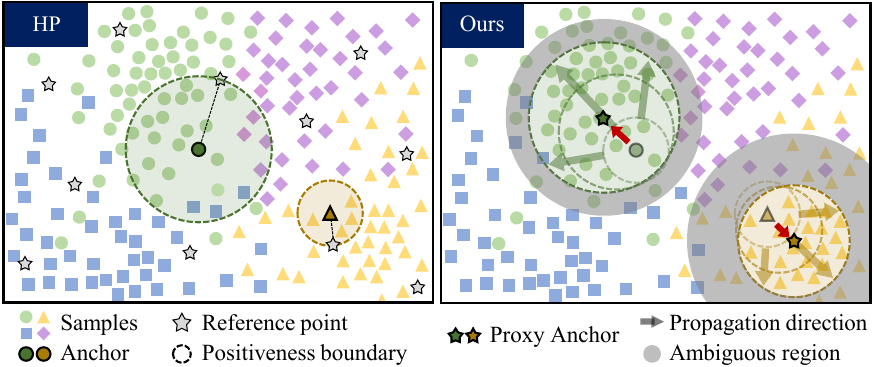}
    \caption{Illustration comparing HP and Ours.}
    \label{fig_fig1_a}
  \end{subfigure}
  \hfill
  \begin{subfigure}{0.29\linewidth}
    % \fbox{\rule{0pt}{0.5in} \rule{.9\linewidth}{0pt}}
    \includegraphics[width=1\textwidth]{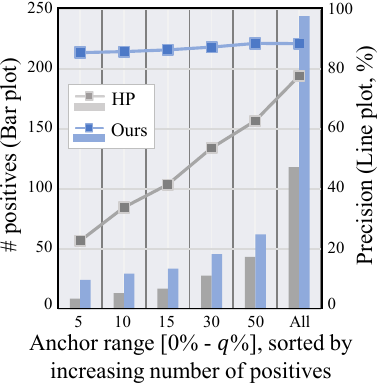}
    \caption{Trustworthiness.}
    \label{fig_fig1_b}
  \end{subfigure}
  \caption{
    (a)~Illustration of how positive and negative sets are determined in HP~\cite{hp} and Ours. 
    Different colors indicate different potential classes.
    In HP, $k$-th nearest neighbor on a per-sample basis becomes an instance-wise positiveness criterion, and all other samples farther than   $k$-th neighbor are considered as the negatives.
    On the other hand, we progressively propagate the proxy anchor to be relocated in the region surrounded by semantically similar samples.
    Consequently, we enable the trustworthy positive collection with numerous samples in dense regions. 
    Additionally, we define an ambiguous region around the positiveness boundary where semantic boundaries might be ambiguous. By excluding the samples in the ambiguous region in training, we avoid the undesired repulsion between the anchor and the possibly \textit{false positives} in the negative set.
    (b)~The number of positives and their precision with respect to the ground truth label for a randomly sampled subset of the dataset. 
    The X-axis represents anchors within the range of [0\%, $q$\%] in the anchor list, sorted in ascending order by the number of identified positives. 
    The bar plot displays the average number of gathered positives, and the line plot illustrates their precision, as determined by the ground truth labels of both the anchor and the positives.
    }
  \label{fig_fig1}
\end{figure}

% \begin{figure}
%     \centering
%     {\includegraphics[width=0.98\textwidth]{fig/fig1_hp_vs_ours_modified2_v4.pdf} }
% \label{fig_fig1111}
% \end{figure}

The main challenge in USS stems from the lack of supervision to train the model.
To overcome this, prior works~\cite{stego, hp, transfgu} suggested first learning the image-level representation space and then leveraging this knowledge to develop the ability of pixel-level understanding. Likewise, utilizing self-supervised pretrained models~\cite{moco, swav, dino} to provide supervision in USS became mainstream. By employing these foundation models, previous techniques have demonstrated promising results, particularly by learning the relationship among image patches in the dataset~\cite{hp, stego, transfgu}.

Yet, we notice that existing methods still encounter challenges in discovering trustworthy relationships between patches.
For instance, HP~\cite{hp}, the latest USS technique based on contrastive learning, utilized $k$-th nearest neighbor of each anchor to determine an appropriate boundary for positive set selection.
While discretizing samples with such a boundary provides an intuitive basis, it often leads to unreliable supervision.
This is because they exclusively rely on the similarity metric on a per-patch basis within an imperfect embedding learned in an unsupervised manner at the image-level~\cite{dino}.
Consequently, as illustrated in Fig.~\ref{fig_fig1_a}, this approach may cause anchors to gather \textit{false positives}~(\textit{FP})~\footnote{The italic \textit{`false positives~(FP)'} and \textit{`true positives~(TP)'}~\cite{wiki:FPFN} represent samples that incorrectly and correctly included in the set, respectively, throughout the paper.} in the positive set especially when they are located in data-sparse areas or near the semantic boundaries while encouraging anchors in dense regions to repel \textit{FP} in the negative set.
Specifically, in Fig.~\ref{fig_fig1_b}, we present a quantitative comparison between HP and our method, focusing on the number and precision of collected positives.
As observed, we note that only 77.5\% of the positives identified by HP match the ground truth labels, even though the number of positives is insufficient.
Precision further decreases for instances in data-sparse regions; samples with a small number of gathered positives exhibit even lower precision~(bottom 10\% samples retain only 33.81\% precision on discovered positives).
This indicates the inclusion of a substantial number of \textit{FP} in the positive set, thereby attracting semantically dissimilar samples and leading to unstable learning.

To mitigate these issues, we propose a Progressive Proxy Anchor Propagation~(PPAP) strategy to deal with the vulnerability of the per-patch-based similarity metric in an image-level pretrained embedding space.
Our goal is to establish a reliable proxy anchor by considering the data distribution surrounding each anchor, thereby gathering patches with more trustworthy positive and negative relationships minimizing ambiguity.
This approach can also obtain a larger number of training guidance as it enhances the precision of gathered relationships.
Specifically, to discover the position of the proxy anchor, we begin by defining a tight boundary around each anchor to construct a small, reliable positive set.
% which their averaged point becomes an initial proxy point.
% The rationale behind the tight boundary is that the samples within highly adjacent regions are highly likely to share the same semantics even in the image-level pretrained embedding space.
The rationale behind establishing a tight boundary is rooted in the observation that samples within closely adjacent regions are highly likely to share similar semantics even within the image-level pretrained embedding space.
Subsequently, we iteratively undertake the following two steps to enlarge a trustworthy positive set per anchor:
1) Re-define the position of a proxy anchor based on the distribution of identified positive samples, 
2) Lower the similarity threshold for the positiveness criterion, \textit{i.e.}, expand the boundary, according to the reliability of the new proxy anchor position, and gather the updated positive set.
Likewise, by discovering the samples with similar semantics and moving the proxy anchor towards the center point of such samples, we expect the assembly of trustworthy positives.
This strategy enables collecting a large number of positive patches with high precision, as shown in Fig~\ref{fig_fig1_b}.
% Still, although the progressive anchor propagation algorithm enables the assembly of trustworthy positives, a positive boundary might not be a perfect measure to detect all the positive samples.
Still, a positiveness boundary might not be a perfect measure to detect all the positive samples.
In other words,
% we expect the ambiguity of positiveness around the positiveness boundary that may contain a mixture of positive and negative instances.
there exists a degree of ambiguity around the boundary, where both positive and negative instances might coexist.
To address this, we expand the original binary relationship categorization of samples, \textit{i.e.}, positive and negative, for contrastive learning into tri-partite groups, \textit{i.e.}, positive, negative, and ambiguous.
The size of the ambiguous set is determined based on the reliability of the relocated proxy anchor.
Consequently, while utilizing the positive and negative sets in contrastive learning, we disregard the ambiguous set, as including \textit{FP} in the negative set often disrupts the stable training~\cite{incrementalFN}.

Overall, our contributions are summarized as follows:
\begin{itemize}
\item{
    % We propose progressive anchor propagation to ensure the credibility of collected positive sets.
    % By discovering the positive samples and using the averaged positive prototype as a proxy for an anchor, we address the vulnerability of calculating the similarity on a per-sample basis within the embedding space trained in an unsupervised manner.
    We propose Progressive Proxy Anchor Propagation~(PPAP), which systematically gathers trustworthy positive samples for each anchor by progressively analyzing the distribution of the positive samples.
    % It addresses the vulnerability of per-sample basis positive collection methods in USS to its precision.
}
% \item{
% To prevent gathering false positives from k-nn strategy of the previous study, we progressively discover trustworthy space of the embedding space for each anchor, thereby securing more reliable positives.
% }
\item{
    % \textcolor{red}{
    We establish an ambiguity-excluded negative set based on the propagated proxy anchor, defining a semantically ambiguous zone for each anchor. This approach effectively eliminates potential \textit{FP} in the negative set.
    % }
    % By forming an ambiguity-excluded negative set for each anchor, we can effectively eliminate the potential \textit{FP} in the negative set.
    % To the best of our knowledge, the concept of semantically ambiguous zone is introduced for the first time in this paper.
    % We design instance-adaptive semantically ambiguous zones to eliminate the possible false positive samples in the negative set.
    % 
    % To overcome the vulnerability of traditional contrastive loss to false negatives, we adopt a tripartite group of samples into positive, negative, and ignore categories during training.
}
\item{
    The efficacy of our trustworthy contrastive learning is validated by achieving new state-of-the-art performances across diverse datasets.
    % Our simple but intuitive approaches provide a core algorithm for forming trustworthy contrastive learning and its effectiveness is validated via establishing new state-of-the-art performances on multiple experimental results.
    % approach has been validated through experimental results, showcasing state-of-the-art performance in this domain.
}
\end{itemize}

\section{Related Work}

\subsection{Unsupervised Semantic Segmentation}
% Semantic segmentation is a challenging but practical problem with a wide range of applications, \textit{e.g.}, autonomous driving, medical imaging, robotics, and augmented reality.
Semantic segmentation aims to classify the semantics of individual pixels within an image~\cite{dilated, chen2017deeplab, atrous, pyramidpool1, pyramidpool2, cnnseg1, cnnseg2, cnnseg3, park2024task}.
% In recent years, significant progress has been made in semantic segmentation through various techniques~\cite{dilated, chen2017deeplab, atrous, pyramidpool1, pyramidpool2, cnnseg1, cnnseg2, cnnseg3}.
% ; dilated convolution~\cite{dilated}, atrous convolution~\cite{chen2017deeplab, atrous} and pyramid pooling~\cite{pyramidpool1, pyramidpool2} have been proposed to address the small receptive fields in fully convolution networks~\cite{cnnseg1, cnnseg2, cnnseg3}.
In recent years, the integration of transformers into semantic segmentation has emerged as a promising research direction~\cite{maskformer, segformer, pasegformer}.
However, achieving pixel-wise supervision requires extensive human labor.
The necessity of learning semantic segmentation without supervision has become apparent in recent literature~\cite{iic, picie, transfgu, hp, seitzer2023bridging, eagle, depthg, equss}.
Earlier trials~\cite{iic, picie}
% extracted paired features per-patch using data augmentations, then 
learned to maintain consistent semantics across the paired features.
In contrast, recent techniques~\cite{transfgu, stego, hp} have employed Vision Transformer~(ViT) models trained in a self-supervised manner as backbone networks to transfer knowledge to the segmentation head. 
For instance, transFGU~\cite{transfgu} grouped target datasets based on prior knowledge and generated pseudo-labels to train the segmentation model.
STEGO~\cite{stego} tried to maintain the patch relationships in the segmentation head by distilling feature correspondences to segmentation correspondences.
HP~\cite{hp}, on the other hand, identified hidden positives using the $k$-th nearest neighbor criterion to guide the contrastive objective.
Our goal aligns with previous works in seeking pseudo-supervision by considering patch relationships.
However, the key difference lies in our approach of considering data distribution to find trustworthy pseudo-supervision within the imperfect embedding space.
% However, the main difference is that we take the reliability of the gathered relationships with consideration of the data distribution around each anchor.
% \textcolor{red}{SAM 넣을지말지 고민}

\subsection{Self-supervised Representation Learning}
Self-supervised representation learning has long been spotlighted for its effectiveness in providing a decent initialization point for various downstream tasks~\cite{dino, stego}.
There are several prevalent approaches in this domain, including pretext tasks which learn the representation by reconstructing the original input from augmented images~\cite{jigsaw, larsson2016learning, zhang2016colorful, rotnet, dosovitskiy2014discriminative, noroozi2018boosting, chen2021jigsaw}, relation-based approaches~\cite{simclr, moco, swav, byol, pirl} and masked-modeling approaches~\cite{he2022masked, xie2023masked, shi2022adversarial, xie2022simmim}.
While masked-modeling approaches excel at preserving local context, they are often less efficient in learning discriminative representations~\cite{he2022masked, xie2022simmim, huang2023contrastive}.
Therefore, relation-based approaches~\cite{dino, selfpatch}, particularly DINO~\cite{dino}, are popularly employed in the realm of USS~\cite{stego, transfgu, hp}.
% Above all, DINO~\cite{dino} is known to demonstrate good semantic representation among the unsupervised feature extractors.
% Nevertheless, a more detailed analysis of DINO~\cite{dino} depicted in Fig.~\ref{fig_fig1_b} reveals that it does not hold true at the patch-level.
% Therefore, we have developed algorithms to reassign per-pixel semantic clusters, especially for embedding regions where DINO is poor at, and demonstrated excellent advancements of semantic segmentation capability over the existing works utilizing DINO features.
% In this context, we develop algorithms to complement the representation space of DINO at the pixel level by reassigning pixels with consistent semantic clusters.
% In this context, we develop algorithms to complement the representation space of DINO at the patch-level by reassigning per-anchor semantic cluster.
Although the features from DINO are powerful in describing semantics for the whole image, their direct use for semantic segmentation proves effective due to the model being trained at the image-level, as shown in Fig~\ref{fig_fig1_b}. In this regard, we have developed algorithms to complement the representation of DINO for the promotion of such features to reflect pixel-level semantics.

% while their pretrained weights are commonly utilized~\cite{stego, transfgu, hp}, the contrastive objective took place as the main objective to learn the pixel-wise semantics~\cite{iic, picie, hp, stego, maskcontrast}.

% Unsupervised learning 생긴 이유
% With the aim of building intermediate representations that can adapt to various tasks, self-supervised learning has gained the popularity~\cite{jigsaw}. 
% Traditional approaches were to perturb the visual input and inverse..
% 학습하기 위해서 나온 pretex task
% 이후 contrastive..., mae
% 
% 효과성 - 다양한 sup task / unsup task.
% USS에서도 사용됨.
% 우리도 그런데 이걸사용하기 위한 sample에 집중.

\section{Method}

\begin{figure}[t]
    \centering
    {\includegraphics[width=0.99\textwidth]{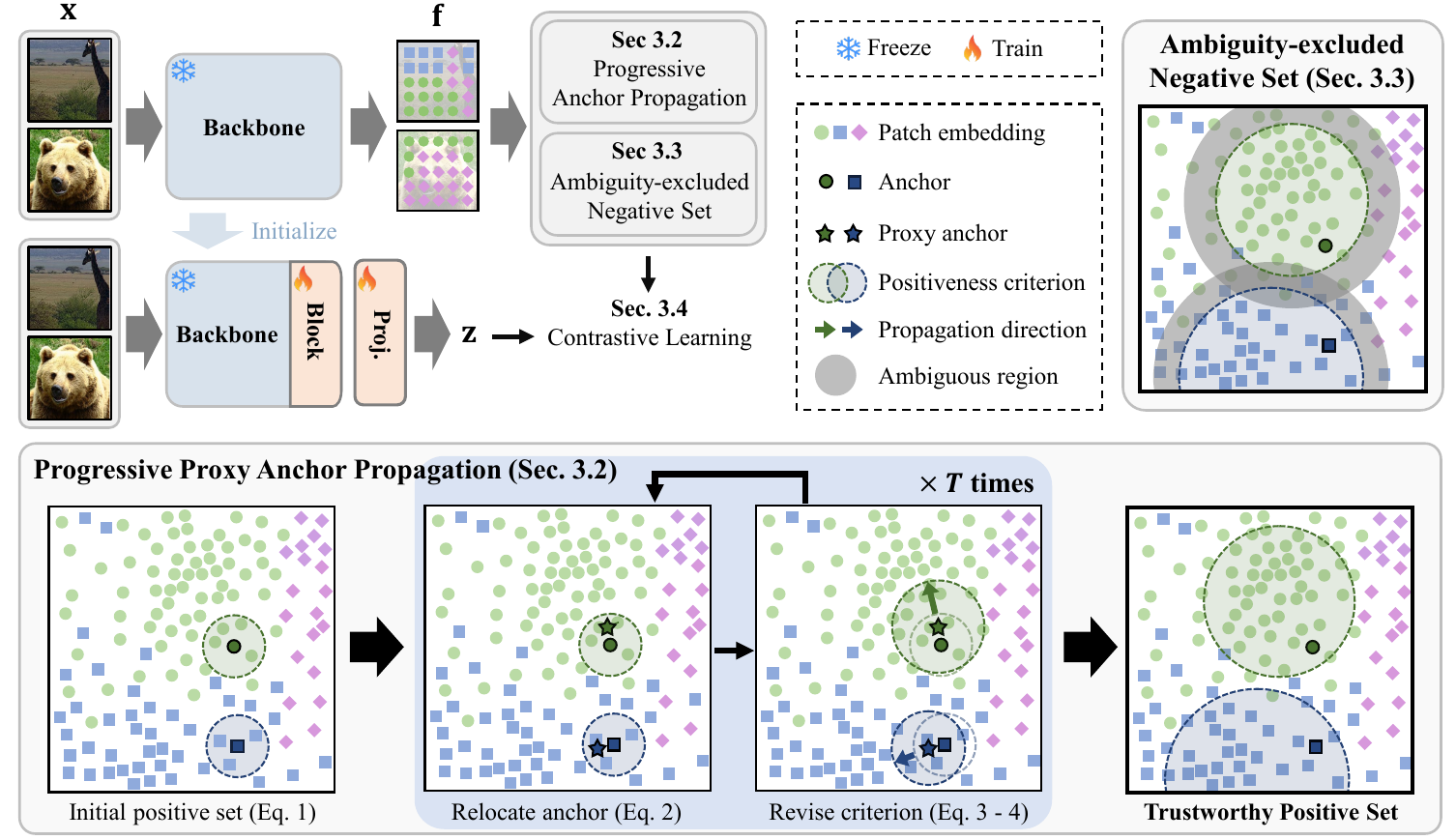} }
    \caption{
    Overall procedure of Progressive Proxy Anchor Propagation~(PPAP). 
    Our backbone consists of two branches: one for acquiring the training guidance, and the other for task adaptive finetuning.
    % While the parameters of the former feature extractor are frozen for stable guidance, we update its last block and the projection head in the latter branch.
    % Employing the pretrained backbone, the last transformer block is copied into two branches: frozen block to provide criterion and finetuning block to adapt to the downstream task.
    % While the feature $\mathbf{f}$ is used to compute the training guidance via trustworthy positive set and ambiguity-excluded negative set from PPAP, the task-adaptive feature $\mathbf{z}$ learns with the obtained training guidance to finetune the task-adaptive branch.
    Specifically, the former feature extractor produces feature $\mathbf{f}$ used to compute training guidance via trustworthy positive and ambiguity-excluded negative sets by PPAP, and its parameters are frozen for stable guidance. On the other hand, the latter branch is being finetuned with the training guidance to learn task-adaptive feature $\mathbf{z}$.
    % An image $x$ is proceeded through each branch to yield $f$ and $g$, respectively.
    % Then, $f$ is utilized to yield training guidance with two 
    % Overall flow of our model. 
    % Our goal is to finetune the last layer of the pretrained ViT backbone to suit the USS task.
    % The feature $\mathbf{f}$, extracted from the frozen backbone, offers criteria tailored for each individual sample. This aids in differentiating between positive set~(Sec.~\ref{sec_progressive_anchor_propagation}) and negative set~(\textit{i.e.}, ambiguous set, Sec.~\ref{sec_ambiguous_set}).
    % Then the last layer and the projection layer are trained using contrastive learning for projected vector $\mathbf{z}$ and the corresponding criteria.
    }
\label{fig_model_overview}
\end{figure}

% \begin{figure*}[t]
%     \centering
%     {\includegraphics[width=0.98\textwidth]{fig/fig_positive_propagation_v2.pdf} }
%     \caption{Illustration of Progressive Anchor Propagation.}
% \label{fig_positive_propagation}
% \end{figure*}

\subsection{Background and Overview} % Preliminary
Recently, it has become mainstream to utilize the positive relationships among patches in training for Unsupervised Semantic Segmentation~(USS)~\cite{stego, transfgu, hp}.
They exploited the patch-wise embeddings from a pretrained foundation model.
However, we claim that they heavily relied on the similarity measured in imperfect embedding space for inferring patch-level training guidance.
% \textcolor{red}{ 
% However, we claim that their training guidance heavily relied on the similarity defined by pretrained foundation model.
% }
Instead, in this paper, we suggest the importance of considering the data distribution; since not all anchors are highly likely to be densely surrounded by semantically similar patch features in the embedding space, we aim to search for better spots to gather a sufficient number of trustworthy positive and negative samples.

The architecture of our method is illustrated in Fig~\ref{fig_model_overview}.
Following the recent works~\cite{stego, transfgu, hp}, our goal is to learn an appropriate projection function for the features extracted from a pretrained model suitable to the USS task.
% To achieve this, we 
To achieve this, we define two streams using the pretrained ViT; for the first stream we keep all the blocks frozen to provide reliable supervision, and for the other stream we finetune the last block for adapting features to the semantic segmentation task.
% To achieve this, we fine-tune the last block of the Vision Transformer~(ViT) model~\cite{vit}, which is pretrained in a self-supervised manner~\cite{dino}.
% Our model overview is illustrated in Fig.~\ref{fig_model_overview}.
% We initially duplicate the last block of DINO; one is a frozen block for defining a criterion and another is an adaptation block for training the task.
Given a mini-batch of images $\{\mathbf{x}_b\}_{b=1}^{B}$, the former stream computes pairs of $B\times H\times W$ patch features $\mathbf{f}_i \in \mathbb{R}^{D}$ where $H\times W$ is the number of patch features for an image and $D$ stands for the dimension of embedding space.
On the other hand, the latter stream producing projected patch features $\mathbf{z}_i \in \mathbb{R}^{D}$ is being finetuned with the gathered positive and negative sets. Specifically, the process begins by determining the positive set $\mathcal{P}_i$ with $\mathbf{f}_i$.
Through the iterative process of positive gathering and proxy anchor relocation, we construct trustworthy positive set.
% Afterward, we determine the ambiguity-excluded negative set to train the task-adaptive feature $\mathbf{z}$. 
Afterward, we determine the ambiguity-excluded negative set to train $\mathbf{z}_i$. 
In the following sections, we discuss the Progressive Proxy Anchor Propagation strategy to obtain trustworthy positive and ambiguity-excluded negative sets.
in Sec.~\ref{sec_progressive_anchor_propagation} and Sec.~\ref{sec_ambiguous_set}, respectively.

\subsection{Progressive Proxy Anchor Propagation}
\label{sec_progressive_anchor_propagation}

% \begin{figure}[t]
%     \centering
%     {\includegraphics[width=0.48\textwidth]{fig/fig_positive_propagation_1col_v5.pdf} }
%     \caption{
%     Illustration of Progressive Anchor Propagation. The process for each anchor $\mathbf{f}_i$ begins by establishing the initial positive set $\mathcal{P}^0_i$ using a tight initial positiveness criterion $\Phi^0_i$. The procedure then involves two following steps.
%     (1) We relocate each anchor towards a more crowded region with semantically-alike samples utilizing the previous positive set~(Eq.~\ref{eq_prototype}). (2) We expand the positive set based on the relocated anchor and revised criterion~(Eq.~\ref{eq_criterion_pos} - \ref{eq_propagated_positive_set}).
%     By moving towards the crowded regions with the same semantics, we expect the anchors to acquire trustworthy positive set $\mathcal{P}_i^T$ with numerous instances from dense regions.
%     % Completing these $T$ steps culminates in acquiring a more trustworthy positive set $\mathcal{P}_i^T$.
%     }
% \label{fig_positive_propagation}
% \end{figure}

% Original
% initial positive를 구했어.
% 그다음 두가지 반복을 할거야.
% prototype 선언.
% positive gahther전에 : adaptive boundary extension
% positive gather
% Newppppppp
% 해야한다.

% 그래서 anchor 주변 positive를 찾고, 그걸로 anchor relocation을 하는 progressive를 할거다.
% 먼저, anchor 주변에 semantic ... 하기 위해 tight.

Collecting a sufficient amount of trustworthy pseudo-supervision is a cumbersome task but crucial for the performance in USS~\cite{incrementalFN}.
% Yet, it is an essential process to ensure the reliability of collected samples for contrastive learning~\cite{incrementalFN}.

% \textcolor{red}{
% To this end, we propose a progressive anchor propagation algorithm with the aim of identifying the dense region that is semantically similar to the given anchor and performing positive sampling in such a region.
% }
% To this end, we propose a progressive anchor propagation algorithm to identify the reliable region for each anchor, where the samples are semantically similar to the given anchor, and to gather trustworthy positive samples, as described in Fig.~\ref{fig_positive_propagation}.
% Briefly, the algorithm employs an iterative strategy composed of two following steps: (1) identify semantically similar positive samples around the anchor, and (2) move the anchor towards more densely populated regions surrounded by the positives. 
To this end, we propose a Progressive Proxy Anchor Propagation algorithm to identify the reliable region for each anchor, where semantically similar samples to each anchor are densely located, as described in Fig.~\ref{fig_model_overview}.
% Briefly, the algorithm employs an iterative strategy composed of two following steps: (1) identify semantically similar positive samples around the anchor, and (2) move the anchor towards more densely populated regions surrounded by the positives. 
% This enables each anchor to gather numerous trustworthy positive samples.
The propagation process begins by forming an initial positive set comprising samples that are highly adjacent to each anchor.
% Subsequently, the algorithm employs an iterative strategy composed of two following steps: (1) relocate the anchor towards more densely populated regions surrounded by the positives, and (2) identify semantically positive samples around the anchor according to the expanded boundary in a density-aware manner.
Subsequently, the algorithm employs an iterative process composed of two following steps: 1) relocate the proxy anchor towards more densely populated regions identified with the distribution of gathered positives, and 2) identify positive samples around the proxy anchor according to the expanded boundary.
% Note that the proxy anchor is the relocated anchor, and the boundary expansion is proportional to the reliability of the proxy anchor's new position that is measured by the propagation degree of the proxy anchor.
Note that the proxy anchor~(\textit{i.e.}, relocated anchor) provides a positive collection criterion on behalf of the anchor where its boundary for positive collection is proportional to the reliability of the proxy anchor’s new position that is measured by the propagation degree of the proxy anchor.
The proxy anchor position is considered more reliable if it does not move significantly, suggesting it is already surrounded by samples with similar semantics.
% We regard the proxy anchor position as more reliable when a proxy anchor does not move far since it infers that the proxy anchor is already surrounded by samples of the same semantics.
This enables each anchor to gather numerous trustworthy positive samples.

Specifically, the initial positive set $\mathcal{P}_i^0$ of a given anchor $\mathbf{f}_i$ is obtained by applying the initial positiveness criterion $\Phi^0$ to gather as below:
% Let us denote the initial positive set $P_i^0$ of an anchor $f_i$ gathered with initial positiveness criterion $c^0$ as below:
\begin{equation}
\label{eq_initial_positive_set}
    \mathcal{P}_i^0 = \{j \mid \mathbf{f}_{i} \cdot \mathbf{f}_{j} > \Phi_i^0, j \in \mathcal{B}\},
    \;\forall_i \Phi_i^0=\Phi^0,
\end{equation}
% \begin{equation}
% \label{eq_initial_positive_set}
%     \mathcal{P}_i^0 = \{j \mid \mathbf{f}_{i} \cdot \mathbf{f}_{j} > \Phi_i^0, j \in \mathcal{B}, \Phi_i^0=\Phi^0 \},
% \end{equation}
where $\mathcal{B}$ denotes the set containing all patch features within the mini-batch, \textit{i.e.}, $|\mathcal{B}|=B\times H\times W$, and $\left(\cdot \right)$ refers to the similarity measure between two vectors~(typically the cosine similarity).
% \textcolor{red}{
% Here, the criterion $\Phi^0$ is to decide whether all other patch features in the mini-batch are positive or not with the similarity threshold that is set to be tight and is shared across all possible anchors.
% Inspiration for the tight boundary of the initial criterion is because samples with high proximity are more likely to be semantically similar which are then used to designate the position for anchor relocation.
% }
Here, the criterion $\Phi^0$ is to decide whether all other patch features in the mini-batch are positive or not, based on the similarity threshold. 
This initial threshold is set to be big enough to make a tight criterion and is shared across all anchors.
Such a tight boundary from a large initial criterion is for stable proxy anchor relocation since the samples with close proximity are more likely to be semantically similar.
% To be specific, the anchor points are progressively propagated to a new position where more reliable positives can be collected

Then, to propagate the proxy anchor toward the positives-dominant region, we derive the new proxy anchor position by averaging the collected positive set.
This way, we take the distribution of the gathered positives into account.
Formally, out of total $T$ steps of relocation, we present the $t$-th relocated position of an anchor $\mathbf{f}_i$ as $\mathbf{v}^t_i$, as follows:
% \begin{equation}
% \label{eq_prototype}
%     \textbf{e}_i^t = \frac{1}{|\mathcal{P}^{t-1}_i|} \sum_{j \in \mathcal{P}^{t-1}_i} p_j.
% \end{equation}
\begin{equation}
\label{eq_prototype}
    \mathbf{v}_i^t = \frac{1}{|\mathcal{P}^{t-1}_i|} \sum_{j \in \mathcal{P}^{t-1}_i} \mathbf{f}_j.
\end{equation}
% \begin{equation}
% \label{eq_prototype}
%     E_i^t = \frac{\tilde{E}^t_i}{\parallel \tilde{E}^t_i \parallel_2}; \quad \tilde{E}^t_i = \frac{1}{|P^{t-1}_i|} \sum_{j \in P^{t-1}_i} p_j.
% \end{equation}
%%%%%%%%%%%%%%%%%%%%%%%%% v1 %%%%%%%%%%%%%%%%%%%%%%%%%%%%
% We posit that the newly calculated position of an anchor is more likely to be centered around the semantically similar patches.
%%%%%%%%%%%%%%%%%%%%%%%%% v2 %%%%%%%%%%%%%%%%%%%%%%%%%%%%
% The close vicinity of the anchor is likely to have a high possibility of semantically similar patches. Consequently, averaging their positions will move the anchor closer to a region that is more densely populated with semantically similar samples.
%%%%%%%%%%%%%%%%%%%%%%%%%%%%v3%%%%%%%%%%%%%%%%%%%%%%%%%%%%%%%%
To account for using the average points of the positive set, we posit that the close vicinity of the proxy anchor is highly likely to retain the same semantic.
Thus, we claim that the center point of the gathered positives will move the proxy anchor closer to a dense region populated with semantically similar samples.
%%%%%%%%%%%%%%%%%%%%%%%%%%%%%%%%%%%%%%%%%%%%%%%%%%%%%%%%%%%%%%

%%%%%%%%%%%%%%%%%%%%%%%%% v1 %%%%%%%%%%%%%%%%%%%%%%%%%%%%
% Yet, the propagated position may not also be reliable so the positiveness criterion $\Phi_i^t$ should be reliability-adaptive.
%%%%%%%%%%%%%%%%%%%%%%%%% v2 %%%%%%%%%%%%%%%%%%%%%%%%%%%%
% As the proxy anchor shifts towards a more reliable region where semantically similar samples are concentrated
Furthermore, the positiveness criterion $\Phi_i^t$ should be reliability-adaptively adjusted.
%%%%%%%%%%%%%%%%%%%%%%%%%%%%%%%%%%%%%%%%%%%%%%%%%%%%%%%%%%%%%%
We determine the reliability of $\mathbf{v}_i^t$ based on the similarity to the previous proxy anchor position $\mathbf{v}_i^{t-1}$ since we assume that the proxy anchor point is converged to the center point of semantically similar patches if the scope of the propagation in a single step is limited~(\textit{i.e.}, high similarity between $\mathbf{v}_i^{t-1}$ and $\mathbf{v}_i^{t}$).
With such intuition, we revise the criterion $\Phi_i^t$ to be loosened when there is high reliability~($\mathbf{v}_i^t$ and $\mathbf{v}_i^{t-1}$ are in close proximity) on the position $\mathbf{v}_i^t$ as follows:
\begin{equation}
\label{eq_criterion_pos}
    \Phi_i^t = \Phi_i^{t-1} - (1 - (\mathbf{v}_i^{t-1} \cdot \mathbf{v}_i^{t}))/\sigma_{\text{pos}},
\end{equation}
where $\sigma_\text{pos}$ is a coefficient used to prevent excessive reduction of the criterion.
Note that $\mathbf{v}_i^0=\mathbf{f}_i$.
% Note that, we replace $\mathbf{v}_i^{t-1}$ refers to $f_i$ when $t=1$.
Consequently, the positive set at iteration $t$ is expressed with the revised criterion $\Phi_i^t$ with the new proxy anchor point $\mathbf{v}_i^t$ as follows:
\begin{equation}
\label{eq_propagated_positive_set}
    \mathcal{P}_i^t = \{j \mid \mathbf{v}_i^t \cdot \mathbf{f}_{j} > \Phi_i^t, j \in \mathcal{B} \}.
\end{equation}

The process above is iteratively performed~(Eq.~\ref{eq_prototype} - Eq.~\ref{eq_propagated_positive_set}) for $T$ times to discover a reliable zone to sample the positives~$\mathcal{P}_i^T$.

\subsection{Ambiguity-excluded Negative Set}
\label{sec_ambiguous_set}
Along with the importance of gathering trustworthy positive sets for contrastive learning, preserving the reliability of the negative sets is another important factor~\cite{incrementalFN}.
Accordingly, we utilize the propagated proxy anchor $\mathbf{v}_i^T$ as the base to compose the negative set to prevent the conflict to the positive set $\mathcal{P}_i^T$.

%%%%%%%%%%%%%%%%%%%%%%%%%%%%%%%%%%%%%%%%%%%%
%%%%%%%%%%%%%%%%%% WJ v1 %%%%%%%%%%%%%%%%%%
% However, we point out the existence of semantic boundaries where semantically ambiguous samples reside.
% Boundaries are particularly ambiguous within the embedding space of dense prediction tasks since patch-wise features may contain multiple semantics in a single patch.
%%%%%%%%%%%%%%%%%%%%%%%%%%%%%%%%%%%%%%%%%%%%
%%%%%%%%%%%%%%%%%%% HS v1 %%%%%%%%%%%%%%%%%%
% However, we point out the presence of an ambiguous zone for each anchor where there is uncertainty in clearly categorizing the samples strictly as positive or negative.
% Derived from such inspiration, we additionally organize the ambiguous set that is neither included in the positive set nor the negative set.
%%%%%%%%%%%%%%%%%%%%%%%%%%%%%%%%%%%%%%%%%%%%
%%%%%%%%%%%%%%%%%%% HS v2 %%%%%%%%%%%%%%%%%%
However, we point out the presence of an ambiguous zone for each anchor where positives and negatives are intermixed, making it unclear to categorize them exactly on one side. 
When the samples in such a zone are considered negatives, the model may face unwanted repulsion. 
Derived from such motivation, we additionally define an ambiguous set for each anchor that is neither included in the positive set nor the negative set, thereby excluding them from the learning process.
%%%%%%%%%%%%%%%%%%%%%%%%%%%%%%%%%%%%%%%%%%%%

The method to establish the ambiguous set is similar to the process for positive set sampling: we update the criterion over the $T$ steps of the proxy anchor propagation in an anchor-dependent manner and define the set according to this criterion.
However, the key difference lies in how we determine the initial ambiguity criterion $\Psi^0$. 
Unlike the boundary $\Phi$ for positive selection, we set $\Psi$ to a small value at the initial step to serve as a loose boundary since the vicinal areas of an initial anchor $\mathbf{f}_i$ might not be reliable.
This criterion is then progressively raised in the subsequent steps.
% But the difference to the formula for positiveness criterion in Eq~\ref{eq_criterion_pos} is that the initial ambiguity criterion $\Psi^0$ is set low, serving a loose boundary at the start. This criterion is then gradually increased throughout the subsequent steps.

Given the initial anchor point $\mathbf{v}_i^{0}$ as $\mathbf{f}_i$ and the $t$-th propagated proxy anchor point $\mathbf{v}_i^t$ through proxy anchor propagation~(Eq.~\ref{eq_prototype}), we progressively adjust the ambiguity criterion by:
% we progressively adjust the area for sampling the ambiguous set by:
% \begin{equation}
% \label{eq_criterion_amb}
%     \Psi_i^t = \Psi_i^{t-1} + \sigma_{\text{amb}}(1 - (\mathbf{e}_i^{t-1} \cdot \mathbf{e}_i^{t})),
% \end{equation}
\begin{equation}
\label{eq_criterion_amb}
    \Psi_i^t = \Psi_i^{t-1} + (1 - (\mathbf{v}_i^{t-1} \cdot \mathbf{v}_i^{t}))/\sigma_{\text{amb}},
\end{equation}
where $\sigma_{\text{amb}}$ is a coefficient to prevent excessive increase of the criterion and $\forall_i\Psi_i^0=\Psi^0$.
% The difference to the formula for positive sampling in Eq.~\ref{eq_criterion_pos} is that we initially set $\Psi^0$ low to define the loose boundary and
Through $t$ steps, $\Psi_i^t$ tightens the boundary if the position of the proxy anchor becomes densely surrounded by the positives. In other words, if the relocated proxy anchor is positioned in a densely populated area, there is less probability of having semantically alike samples outside the positive sampling region formed with Eq.~\ref{eq_criterion_pos}.

After determining the ambiguity criterion through $T$ steps, we then proceed to define the ambiguous set.
% Upon completing $T$ steps, we determine the criterion for the ambiguous set.
Using the $T$-th relocated proxy anchor $\mathbf{v}_i^T$, the ambiguous set $\mathcal{A}_i$ for the anchor $\mathbf{f}_i$ is defined as follows:
\begin{equation}
\label{eq_ambiguous_set}
    \mathcal{A}_i = \{j \mid (\mathbf{v}_i \cdot \mathbf{f}_j > \Psi_i^T) \land (\mathbf{v}_i \cdot \mathbf{f}_j < \Phi_i^T), j \in \mathcal{B} \}.
\end{equation}
Finally, the negative set $\mathcal{N}$ is organized as follows:
\begin{equation}
\label{eq_negative_set}
    \mathcal{N}_i = \{j \mid (j \notin \mathcal{P}_i) \land (j \notin \mathcal{A}_i), j \in \mathcal{B}\}.
\end{equation}

% Finally, the negative set $\mathcal{N}_i$ is organized as follows:
% \begin{equation}
% \label{eq_propagated_negative_set}
%     \mathcal{N}_i = \{j \mid 
%     \textbf{e}_i^t \cdot \textbf{f}_{j} < \Psi_i^t, j \in \mathcal{B} \}.
% \end{equation}

\subsection{Training Objective}
Following existing works~\cite{picie, hp, iic, stego} in USS, we utilize contrastive learning objecitve~\cite{simclr, supcon}.
With the aim of distinguishing the semantically similar positive set $\mathcal{P}_i^T$ and dissimilar negative set $\mathcal{N}_i$, the objective is expressed as:
\begin{equation}
\label{eq_contrastive_loss}
    L^{\text{con}}_i =
    \frac{-1}{|\mathcal{P}^T_i|}\sum_{p \in \mathcal{P}^T_i} \text{log} \frac{\text{exp}(\mathbf{z}_i \cdot \mathbf{z}_p / \tau)}{\sum\limits_{n \in (\mathcal{N}_i \cup \mathcal{P}_i^T)} \text{exp}(\mathbf{z}_i \cdot \mathbf{z}_n / \tau)},
\end{equation}
where $\tau$ is a temperature parameter, and $\mathcal{P}^T_i$ and $\mathcal{N}_i$ denote positive and negative sets for $i$-th anchor, respectively.
\section{Experiments}

\subsection{Experimental Settings}
\subsubsection{Datasets.}
% Following previous protocols~\cite{iic, picie, stego, hp}, we evaluate our method on COCO-stuff~\cite{coco} and Cityscapes~\cite{cityscapes} datasets.
Following previous protocols~\cite{iic, picie, stego, hp, INS}, we evaluate our method on COCO-stuff~\cite{coco} and Cityscapes~\cite{cityscapes}, Potsdam-3, and ImageNet-S~\cite{INS} datasets. Further details can be found in the Appendix.
COCO-stuff is a dataset for scene understanding tasks, \textit{e.g.}, semantic segmentation, detection, and image captioning, that consists of 172 classes.
Among them, the COCO-stuff benchmark for USS utilizes 27 classes.
Cityscapes is another large-scale dataset for scene understanding that consists of 30 classes captured across 50 different cities.
Similarly to COCO-stuff, 27 subclasses are used for the benchmark.
In addition, Potsdam-3 contains satellite images that are divided into 3 classes.
Lastly, ImageNet-S is a large-scale dataset which has 1.2 million training images with 919 semantic classes.

\subsubsection{Evaluation Protocols.}
% To compare between the methods,
For COCO-stuff, Cityscapes, and Potsdam-3 datasets, we adopt two evaluation methods: clustering~(unsupervised) and linear probe~\cite{stego, hp}.
Clustering evaluates the alignment between the prediction and the ground truth with the Hungarian matching algorithm.
On the other hand, the linear probe utilizes an additional fully connected layer for classification.
For both evaluations, we apply the post-processing step using a Conditional Random Field~(CRF)~\cite{crf} to refine the predictions.
Accuracy~(Acc.) and mean Intersection over Union~(mIoU) are used to measure the performances.
For the evaluation on the ImageNet-S dataset, we adopt mIoU with distance matching~\textit{i.e.}, the k-nearest neighbors classifier with k=10, following the evaluation protocol from PASS~\cite{INS}.

% \subsubsection{Implementation Details.}
% In line with existing works~\cite{stego, hp, transfgu}, we utilize a DINO-pretrained ViT as our backbone network. The embedding dimension of the projection layer~($D$) is 384 for ViT-small and 768 for ViT-base.
% % The initial positiveness and ambiguity criteria ($\Phi^0$, $\Psi^0$) is set as (0.55, 0.2) for ViT-S/8, (0.55, 0.15) for ViT-S/16, and (0.6, 0.2) for ViT-B/8 model.
% We set $T$ to be 2, 3 and 1 for the experimental results in Tab.~\ref{table_coco},~\ref{table_cityscapes} and~\ref{table_potsdam}, respectively.
% % \textcolor{red}{The initial positiveness criterion is set between 0.55 and 0.6, while the initial ambiguity criterion is set in the range of 0.15 to 0.2 for all experiments.
% % More details are in Appendix.
% % }
% Detailed hyperparameter settings and discussions are in Sec.~\ref{sec_ablation_study} and Appendix.
% % For all our experiments, we maintain $\sigma$ at 3 to ensure a stable positive propagation.

% Main table
\begingroup
\setlength{\tabcolsep}{3pt} % Default value: 6pt
\renewcommand{\arraystretch}{1.0} % Default value: 1
\begin{table}[t!]
    \centering
    \caption{Experimental results on COCO-stuff dataset.}
    % \small
    \scriptsize
    \begin{tabular}{l | c| c c | c c}
        \toprule
        \multirow{2}{*}{Method}& \multirow{2}{*}{Backbone} & \multicolumn{2}{c|}{Unsupervised} & \multicolumn{2}{c}{Linear} \\
        & & Acc. & mIoU & Acc. & mIoU \\
        \midrule
        DC~\cite{deepcluster} & R18+FPN & 19.9 & - & - & - \\
        MDC~\cite{deepcluster} & R18+FPN & 32.2 & 9.8 & 48.6 & 13.3 \\
        IIC~\cite{iic} & R18+FPN & 21.8 & 6.7 & 44.5 & 8.4 \\
        PiCIE~\cite{picie} & R18+FPN & 48.1 & 13.8 & 54.2 & 13.9 \\
        PiCIE+H~\cite{picie} & R18+FPN & 50.0 & 14.4 & 54.8 & 14.8 \\
        \midrule
        DINO~\cite{dino} & ViT-S/8 & 28.7 & 11.3 & 68.6 & 33.9 \\ 
        TransFGU~\cite{transfgu} & ViT-S/8 & 52.7 & 17.5 & - & - \\
        STEGO~\cite{stego} & ViT-S/8 & 48.3 & 24.5 & 74.4 & 38.3 \\
        HP~\cite{hp} & ViT-S/8 & 57.2 & 24.6 & 75.6 & 42.7 \\ 
        \rowcolor{gray!30}
        PPAP~(Ours) & ViT-S/8 & \textbf{59.0} & \textbf{27.2} & \textbf{76.9} & \textbf{46.3} \\ 
        % + Ours v2(변동) & ViT-S/8 & \textbf{61.1} & \textbf{26.0} & \textbf{76.9} & \textbf{44.5} \\ 
        % \hlineB{2.5}
        % \multicolumn{5}{c}{Vit-base} \\
        \midrule
        DINO~\cite{dino} & ViT-S/16 & 22.0 & 8.0 & 50.3 & 18.1 \\ 
        STEGO~\cite{stego} & ViT-S/16 & 52.5 & 23.7 & 70.6 & 34.5 \\ 
        HP~\cite{hp} & ViT-S/16 & 54.5 & 24.3 & 74.1 & 39.1 \\
        % + Ours v1 & ViT-S/16 & \textbf{57.5} & \textbf{26.6} & \textbf{75.8} & \textbf{43.1} \\  % dino16-011 opt5
        % + Ours v2 & ViT-S/16 & \textbf{59.0} & \textbf{26.6} & \textbf{75.0} & \textbf{41.4} \\  % dino16-016
        % + Ours(변동) & ViT-S/16 & \textbf{57.9} & \textbf{25.4} & \textbf{75.9} & \textbf{43.8} \\  % dino16-23lr low0.2 sigma 5
        \rowcolor{gray!30}
        PPAP~(Ours) & ViT-S/16 & \textbf{62.9} & \textbf{26.5} & \textbf{76.0} & \textbf{43.3} \\  % dino16-23 low0.15 sigma 4
        \bottomrule
    \end{tabular}
    \label{table_coco}
    % \vspace{-0.1cm}
    % \vspace{-0.4cm}
\end{table}
\endgroup

\subsection{Experimental Results}
% \paragraph{COCO-stuff dataset}

\subsubsection{Quantitative Result.}
We compare the performances of our method with various baselines~\cite{deepcluster, iic, picie, transfgu, stego, hp}.
In Tab.~\ref{table_coco}, we observe that the recent works with the ViT backbone outperform the other ones, and among them, our approach demonstrates state-of-the-art performances across all metrics.
In particular, our PPAP, equipped only with the sampling strategies for both the positive and negative samples, exceeds HP~\cite{hp} that utilizes contrastive learning also with locality learning in a task-specific perspective.

\begin{table*}[!t]
    \centering
    \small
    \begin{minipage}[t!]{0.53\linewidth}%\centering
    \setlength{\tabcolsep}{1.5pt} % Default value: 6pt
    \renewcommand{\arraystretch}{1.0} % Default value: 1
    \centering
    {
        \caption{Experimental results on Cityscapes dataset.}
        \scriptsize
        \begin{tabular}{l| c| c c | c c}
        \toprule
        \multirow{2}{*}{Method} & \multirow{2}{*}{Backbone} &
        \multicolumn{2}{c|}{Unsupervised} & \multicolumn{2}{c}{Linear} \\
        & & Acc. & mIoU & Acc. & mIoU \\
        \midrule
        MDC~\cite{deepcluster} & R18+FPN & 40.7 & 7.1 & - & - \\
        IIC~\cite{iic} & R18+FPN & 47.9 & 6.4 & - & - \\
        PiCIE~\cite{picie} & R18+FPN & 65.5 & 12.3 & - & - \\
        \midrule
        DINO~\cite{dino} & ViT-S/8 & 34.5 & 10.9 & 84.6 & 22.8 \\
        TransFGU~\cite{transfgu} & ViT-S/8 & 77.9 & 16.8 & - & - \\
        HP~\cite{hp} & ViT-S/8 & 80.1 & 18.4 & \textbf{91.2} & 30.6 \\ % ex129
        \rowcolor{gray!30}
        PPAP~(Ours) & ViT-S/8 & \textbf{82.0} & \textbf{19.6} & 90.8 & \textbf{31.5} \\ % citysmall 29
        % \hline
        % DINO~\cite{dino} & ViT-S/8 & 34.5 & 10.9 & 84.6 & 22.8 \\
        % + TransFGU~\cite{transfgu} & ViT-S/8 & 77.9 & 16.8 & - & - \\
        % + HP & ViT-S/8 & 80.1 & 18.4 & 91.2 & 30.6 \\
        % + Ours & ViT-S/8 &  &  &  &  \\ 
        \midrule
        DINO~\cite{dino} & ViT-B/8 & 43.6 & 11.8 & 84.2 & 23.0 \\
        STEGO~\cite{stego} & ViT-B/8 & 73.2 & 21.0 & 90.3 & 26.8 \\
        HP~\cite{hp} & ViT-B/8 & 79.5 & 18.4 & 90.9 & 33.0 \\ 
        \rowcolor{gray!30}
        PPAP~(Ours) & ViT-B/8 & \textbf{83.3} & \textbf{21.2} & \textbf{91.4} & \textbf{36.5} \\ % city base 17-param
        \bottomrule
        \end{tabular}
        \label{table_cityscapes}
        }
    \end{minipage}\hfill%
    \begin{minipage}[t!]{0.44\linewidth}
    \setlength{\tabcolsep}{2pt} % Default value: 6pt
    \renewcommand{\arraystretch}{1.0} % Default value: 1
    \centering
    {
        \caption{Experimental results on Potsdam-3 dataset.}
        \scriptsize
        \begin{tabular}{l| c| c c c }
        \toprule
        % Method & Backbone & Unsup. Acc. \\
        \multirow{2}{*}{Method} & \multirow{2}{*}{Backbone} &
        \multicolumn{2}{c}{Unsupervised} \\
        & & Acc. & mIoU \\
        \midrule
        Rand.~CNN~\cite{iic}& VGG11 & 38.2 & - \\
        K-Means~\cite{sklearn}& VGG11 & 45.7 & - \\
        SIFT~\cite{sift} & VGG11 &38.2 & - \\
        CP~\cite{context} & VGG11 & 49.6 & - \\
        CC~\cite{cc} & VGG11 & 63.9 & - \\
        DeepCluster~\cite{deepcluster} & VGG11 & 41.7 & -\\
        IIC~\cite{iic} & VGG11 & 65.1 & - \\
        \midrule
        DINO~\cite{dino} & ViT-B/8 & 53.0 & - \\
        STEGO~\cite{stego}& ViT-B/8 & 77.0 & - \\
        HP~\cite{hp} & ViT-B/8 & 82.4 & 69.7 \\
        \rowcolor{gray!30}
        PPAP~(Ours) & ViT-B/8 & \textbf{83.2} & \textbf{71.0} \\ 
        \bottomrule
        \end{tabular}
        \label{table_potsdam}
        }
    \end{minipage}%\hfill
\end{table*}

\begin{table*}[!t]
    \centering
    \small
    \begin{minipage}[t!]{0.53\linewidth}%\centering
    \setlength{\tabcolsep}{3.8pt} % Default value: 6pt
    \renewcommand{\arraystretch}{1.0} % Default value: 1
    \centering
    {
        \caption{Experimental results on ImageNet-S validation set. $\dagger$: reproduced performance.}
        \scriptsize
        \begin{tabular}{l| c | c c c c}
        \toprule
        % Method & Backbone & Unsup. Acc. \\
        Method & Backbone & IN-S & IN-S$_{300}$ & IN-S$_{50}$ \\
        \midrule
        SwAV~\cite{swav} & ResNet-50 & 15.1 & 22.4 & - \\
        PASS~\cite{INS} & ResNet-50 & 15.6 & 25.1 & - \\
        \midrule
        DINO${^\dagger}$~\cite{dino} & ViT-S/16 & 7.3 & 12.0 & 22.8 \\
        HP${^\dagger}$~\cite{hp} & ViT-S/16 & 8.6 & 14.4 & 29.5 \\
        \rowcolor{gray!30}
        PPAP~(Ours) & ViT-S/16 & \textbf{25.7} & \textbf{37.3} & \textbf{59.3} \\
        \bottomrule
        \end{tabular}
        \label{table_IN-S}
    }
    \end{minipage}\hfill%
    \begin{minipage}[t!]{0.44\linewidth}
    \setlength{\tabcolsep}{8pt} % Default value: 6pt
    \renewcommand{\arraystretch}{1.0} % Default value: 1
    \centering
    {\scriptsize
        \caption{Ablation study varying each component.
        % TPS: Trustworthy Positive Set. AeN: Ambiguity-excluded Negative Set.
        }
        % \small
        \begin{tabular}{c |c |c c}
        \toprule
        \multicolumn{2}{c|}{PPAP} & \multicolumn{2}{c}{Unsupervised} \\
        % \cmidrule{0-1}
        TPS & ANS & Acc. & mIoU \\
        \midrule
        -& -& 49.1 & 22.9 \\
        \checkmark & -& 53.7 & 25.0 \\
        -& \checkmark & 55.5 & 25.1 \\
        \checkmark & \checkmark & 62.9 & 26.5 \\
        \bottomrule
        \end{tabular}
        \label{Tab.ablation}
        }
    \end{minipage}%\hfill
\end{table*}

% According to Tab.~\ref{table_coco}, our approach demonstrates state-of-the-art performances across all metrics, notably even when compared with methods based on the DINO pretrained model~\cite{dino}. 
% Among them, our method outperforms those sharing the idea of contrastive learning~\cite{hp, stego}. 
% This suggests that using trustworthy positives gathered by our method in contrastive loss significantly enhances performance.

In terms of the backbone, we achieve greater improvements with the ViT-S/16, which uses larger-sized patch features.
We attribute these results to the robust property of our PPAP.
Unlike PPAP, other methods are shown to yield better results with the small patch size~(ViT-S/8 backbone) since the patches with the larger size are more likely to include a mixture of semantics.
Yet, our proposed PPAP is guided to search for semantically similar patches to learn its prototypical proxy point and even disregard the patches that retain an ambiguous relationship with the given anchor.
As a result, PPAP achieves promising results by a margin up to 15.41\% and 9.05\% compared to HP~\cite{hp} in unsupervised Acc. and mIoU, respectively.

% Patches with mixed semantics are more likely to be located in sparser regions rather than at the semantic center in the embedding space. When such patches serve as anchors, our algorithm efficiently identifies and utilizes nearby spaces with a high probability of having identical semantics as positives. We argue that this aspect leads to greater benefits in our approach.

% \paragraph{Cityscapes dataset}
Performance comparison on the Cityscapes dataset is displayed in Tab.~\ref{table_cityscapes}.
Similar to the results on the COCO-stuff dataset, our proposed PPAP achieves new state-of-the-art results except in one case.
These results further verify the applicability of our components to different datasets.
% In addition, we compare our method with the baselines in another dataset in Tab.~\ref{table_cityscapes}. 
% We can observe that there is a consistent improvement in performance over existing methods on the Cityscapes dataset. 
% These results validate the applicability of our method for gathering trustworthy positives and considering ambiguity.

%PPAP also shows improvement on the Potsdam-3 as shown in Tab.~\ref{table_potsdam},  but it is marginal since the dataset is brief enough 
%to well cluster by pretrained ViT at the image level.
PPAP also shows improvement on the Potsdam-3 as shown in Tab.~\ref{table_potsdam}, but the difference is modest due to the dataset's limited three distinct semantic classes, well clustered by pretrained ViT features. However, our method demonstrates greater enhancements on datasets with more and less distinct semantic classes, not as effectively distinguished by the pretrained backbones.

On the ImageNet-S dataset, PPAP significantly outperforms existing methods, PASS~\cite{INS} and HP~\cite{hp}, as shown in Tab.~\ref{table_IN-S}. We also conduct experiments on the subsets ImageNet-S$_{300}$ and ImageNet-S$_{50}$, which contain 300 and 50 classes, respectively.
The superior performance of PPAP on these datasets further demonstrates its scalability compared to existing methods.

% In the PPAP algorithm, the method of propagating the proxy anchor is similar to the algorithm of searching centers in k-means clustering.
% PPAP 알고리즘에서 proxy anchor를 propagation하는 방식은 평균을 취하고 center를 relocate한다는 점에서 k-means clustering에서 center를 찾아가는 방식과 유사하다.
% 하지만 PPAP는 이에 더 나아가, 매우 인접한 nearest 들을 이용해 anchor의 초기 이동을 안정적으로 수행하고, anchor가 relocate되는 정도에 기반해 각 anchor별 positiveness\&negativeness 를 적절히 정해준다.
% Our proxy anchor propagation algorithm is similar to how k-means clustering finds and relocates the centers through averaging.
% However, PPAP enhances this process by using very close nearest neighbors to stabilize the initial movement of each anchor. It also appropriately determines the positiveness and negativeness of each anchor based on the degree of its relocation.
% Without such a process, simply applying k-means clustering and assigning positive relationships to all features within each cluster results in relatively high recall but low precision in the positive set. For example, we observe that the precision of positives with k-means is only 22\% even when class number given as a prior, while PPAP achieves 72\% precision on the COCO-stuff dataset using DINO pretrained ViT-S/16.
% We have empirically demonstrated that the low precision of a positive set correlates with the performance degradation of the USS task, as shown in Table~\ref{table_coco},~\ref{table_cityscapes}, and~\ref{table_potsdam}~\cite{deepcluster}.
Our PPAP algorithm shares similarities with k-means clustering in that both methods iteratively find and relocate points through averaging.
% but PPAP at the instance~(each patch) level. 
However, there are three critical differences:
1) PPAP aims to discover patch-wise proxy anchor rather than having instances within a cluster share a single proxy.
2) PPAP enables a stable initial relocation process by using only the very close nearest neighbors of an anchor.
3) PPAP accurately determines the positiveness and negativeness of each anchor based on the degree of its relocation.
In contrast, simply applying k-means clustering and assigning positive relationships to all features within each cluster leads to high recall but low precision in the positive set. For example, we observed that the precision of positives with k-means is only 22\%, even when the number of classes is given as a prior. In comparison, PPAP achieves a precision of 72\% on the COCO-stuff dataset using the DINO pretrained ViT-S/16 model without class prior.

%. The main reason of insignificant performance improvements is that the dataset only contains 3 different classes so that each semantic classes are simple enough to well cluster by the pretrained ViT at the image level.
%, but it is not significant since the dataset is brief enough to well cluster by pretrained ViT at the image level.

% \paragraph{Potsdam-3 dataset}

\subsubsection{Qualitative Result.}
\begin{figure}[t]
    \centering
    {\includegraphics[width=0.99\textwidth]{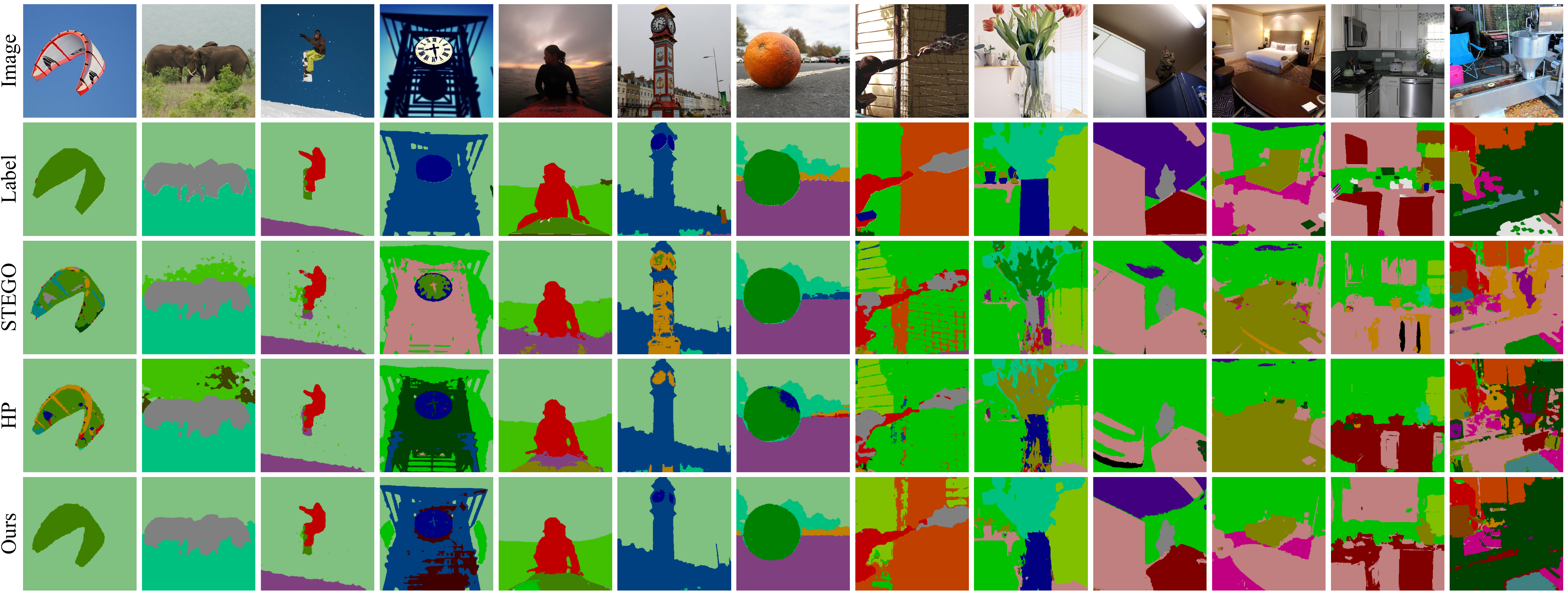}}
    \caption{
    Qualitative comparison results of PPAP~(Ours) with STEGO and HP on the COCO-stuff dataset with DINO pretrained ViT-S/8 backbone.
    }
\label{fig_visualize_seg}
\end{figure}

We display our quantitative results in comparison to STEGO~\cite{stego} and HP~\cite{hp} in \cref{fig_visualize_seg}.
Considering the complexity of the scenes, we plot simple to complicated scenes in order from left to right.
To be brief, our PPAP shows consistent results to have fewer mispredicted pixels compared to the baselines.
Particularly, our proposed method is robust to pixel-wise noises because each anchor is progressively propagated to search for reliable points that address the vulnerability of per-sample basis inference in the imperfect embedding space learned in an unsupervised manner at the image-level.

% Besides the quantitative results, we additionally report qualitative results in \cref{fig_visualize_seg}.
% From the simple scenes~(left) to the complicated scenes~(right), our results show fewer mispredicted pixels than existing works.
% This is because our method reflects the semantics not the color of objects.
% 안녕 나는 세옹이야!
% For instance, when objects~(glider in the 1-st column and clock tower in the 6-st column) consist of various colors, our method preserves their semantic consistency.
% However, existing works show sensitivity to color.
% These results verify our superiority.

\subsection{Ablation Study}
\label{sec_ablation_study}
% 모든 ablation study는 COCO-stuff/16에서 수행한다.
We provide ablation studies to evaluate the individual components and key hyperparameters.
The primary components under study are: 
1) Trustworthy Positive Set~(TPS) obtained based on PPAP and 2) Ambiguity-excluded Negative Set~(ANS).
Additionally, we examined the impact of hyperparameters, specifically: 1) coefficients $\sigma_\text{pos}$ and $\sigma_\text{amg}$ to regulate the criteria and 2) initial criteria $\Phi^0$ and $\Psi^0$.

\subsubsection{Varying Components of PPAP.}
We perform an ablation study to assess the individual contributions of each component, which are presented in Tab.~\ref{Tab.ablation}. 
These experiments are performed on the COCO-stuff dataset with ViT-S/16 backbone.
For the baseline, we train the model using contrastive loss based on positives determined with the initial positiveness criterion $\mathcal{P}^0$ and negatives comprising all remaining samples.
Incorporating the TPS yields improvements of 9.36\% in unsupervised accuracy and 9.17\% in mIoU.
For the experiment on the third row, the ambiguous set for contrastive loss is defined after $T$-step propagation, while the positive set is defined with the initial positiveness criterion $\mathcal{P}^0$.
This brings 13.03\% and 9.6\% enhancements over the baseline.
These results verify that both components significantly contribute to performance enhancement.
Consequently, with both components combined, we observe a notable overall improvement of 28.11\% in accuracy and 15.72\% in mIoU. 
% This substantial enhancement confirms the effectiveness and trustworthiness of positive and negative sets identified by our method.

% \begin{figure*}[t]
%     \centering
%     {\includegraphics[width=0.98\textwidth]{fig/fig_ablation_study_v4.pdf} }
%     \caption{asd}
% \label{fig_ablation}
% \end{figure*}

% \begin{figure}[t]
%     \centering
%     {\includegraphics[width=0.48\textwidth]{fig/fig_ablation_study_1col_v3.pdf} }
%     \caption{Ablation studies for various coefficients.}
% \label{fig_ablation}
% \end{figure}

% \begin{figure}
%     {\includegraphics[width=0.48\textwidth]{fig/fig_ablation_study_sigma_v1.pdf} }
%     \caption{왼쪽: sigma pos / 오른쪽: sigma neg}
% \label{fig_ablation}
% \end{figure}

% \begin{figure}
%     {\includegraphics[width=0.48\textwidth]{fig/fig_ablation_study_sigma_v1.pdf} }
%     \caption{왼쪽: pos initial crit / 오른쪽: neg initial crit}
% \label{fig_ablation}
% \end{figure}

% \begin{figure}
%     {\includegraphics[width=0.48\textwidth]{fig/fig_ablation_study_sigma_v1.pdf} }
%     \caption{왼쪽: T step (COCO) / 오른쪽: T step (City)}
% \label{fig_ablation}
% \end{figure}

% \begin{figure}[t]
%     \centering
%     {\includegraphics[width=0.99\textwidth]{fig/fig_ablation_study_3datasets_v4.pdf} }
%     \caption{Ablation studies of various coefficients on three different datasets. Whereas the X-axis denotes the value of each hyperparameter, the Y-axis shows the performance.}
% \label{fig_ablation}
% \end{figure}

\begin{figure}[t]
  \centering
  \begin{subfigure}{0.24\linewidth}
        \includegraphics[width=1\textwidth]{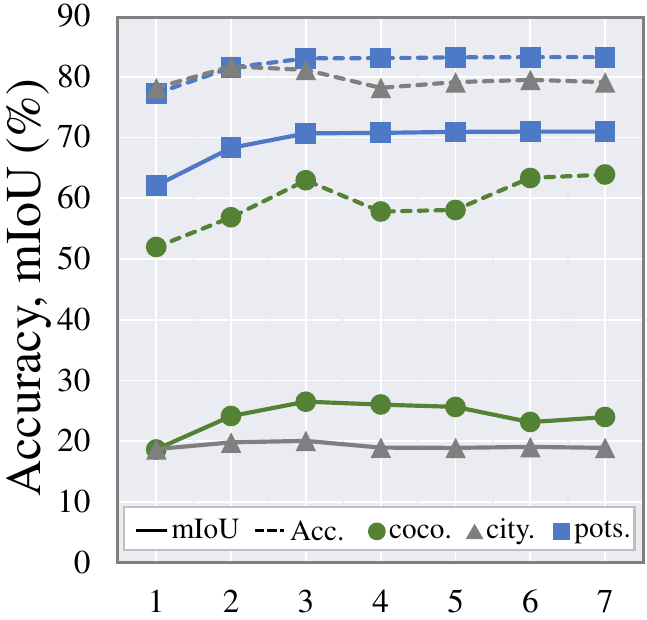}
    \caption{$\sigma_{\text{pos}}$}
    \label{fig_ablation_a}
  \end{subfigure}
  \begin{subfigure}{0.24\linewidth}
    \includegraphics[width=1\textwidth]{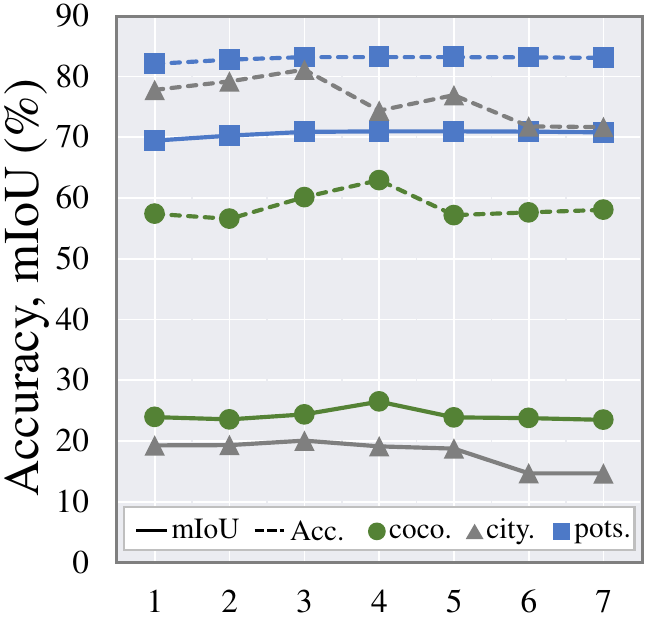}
    \caption{$\sigma_{\text{amb}}$}
    \label{fig_ablation_b}
  \end{subfigure}
  \begin{subfigure}{0.24\linewidth}
        \includegraphics[width=1\textwidth]{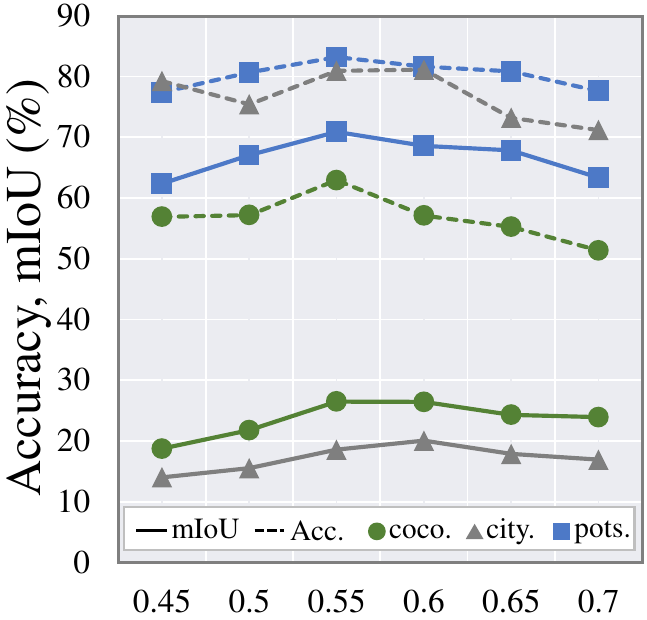}
    \caption{$\Phi^0$}
    \label{fig_ablation_c}
  \end{subfigure}
  \begin{subfigure}{0.24\linewidth}
        \includegraphics[width=1\textwidth]{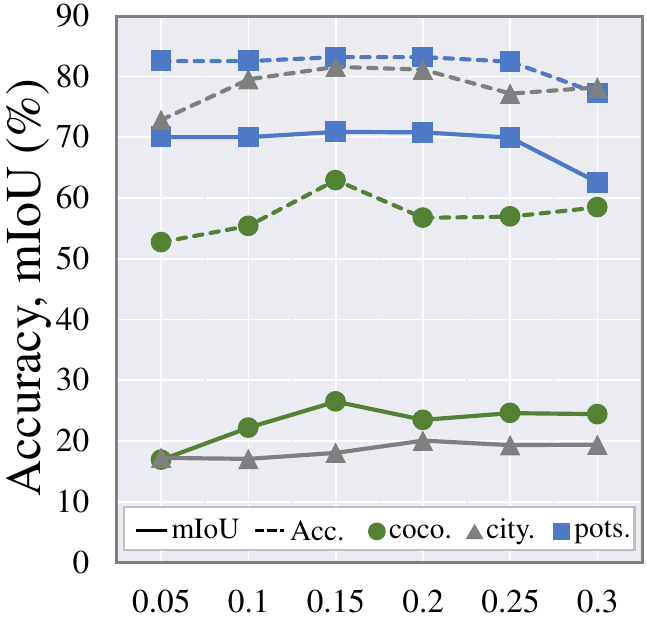}
    \caption{$\Psi^0$}
    \label{fig_ablation_d}
  \end{subfigure}
  \caption{
  Ablation studies of various coefficients on three different datasets. Whereas the X-axis denotes the value of each hyperparameter, the Y-axis shows the performance.
    }
  \label{fig_ablation}
\end{figure}

\subsubsection{Varying $\sigma_\text{pos}$ and $\sigma_\text{amb}$.}
In Fig~\ref{fig_ablation_a} and \ref{fig_ablation_b}, we carry out an ablation study on the coefficient $\sigma$, which is crucial for controlling the degree of reduction and increase in positiveness and ambiguity criteria, respectively.
A smaller $\sigma_\text{pos}$ leads to a more intensive reduction in the positiveness criterion, while a bigger $\sigma_\text{pos}$ results in a more gradual reduction.
Such property is reflected in the outcomes presented in Fig~\ref{fig_ablation_a}. 
A smaller $\sigma_\text{pos}$ tends to lower the performance due to $FP$ erroneously included in the positive set. 
On the other hand, moderately larger $\sigma_\text{pos}$ helps to mitigate the aforementioned problem.
$\sigma_\text{amb}$ follows the same principle: a smaller value results in a more substantial increase of the criterion, escalating the possibility of erroneously repelling $FP$ in the negative set. And a bigger value leads to a more moderate increase.
A proper $\sigma$ can prevent both the excessive changes in the criteria.
We note that it is advisable to choose values around 3 to ensure stability.

\subsubsection{Varying $\Phi^0$ and $\Psi^0$.}
Ablation studies for varying $\Phi^0$ and $\Psi^0$ are shown in Fig.~\ref{fig_ablation_c} and \ref{fig_ablation_d}.
These parameters act as the initial criteria for selecting positive and ambiguous samples and serve as key hyperparameters in our method.
For $\Phi^0$, as samples in close vicinity to the anchor are highly likely to share the same semantics, setting its initial value low incurs the existence of $FP$ in the positive set.
Still, setting the $\Phi^0$ too high results in selecting only a few positives that the anchor relocation is implemented only within its very close proximity.
Regarding $\Psi^0$, setting $\Psi^0$ too small significantly reduces the size of the negative set and leads to a shortage of hard negative samples in the negative set, while too large $\Psi^0$ increases $FP$ in the negative set.
We found that values of 0.55 for $\Phi^0$ and 0.15 for $\Psi^0$ consistently perform well across all datasets and models.
Note that we also found that higher $\Phi^0$ requires more propagation steps~(\textit{i.e.,} bigger $T$).

\begingroup
\setlength{\tabcolsep}{6pt}
\begin{table}[t]
\renewcommand{\arraystretch}{1.0} % Default value: 1
	\centering
    \caption{Comparing both \textit{TP} in positive set~($\mathcal{P}$) and \textit{FP} in negative set~($\mathcal{N}$) between HP and Ours on 3 datasets. Numbers and percentages are the results averaged from subsamples of each dataset. S/8 and B/8 indicate ViT-S/8 and ViT-B/8 backbone, respectively.}
    {
    \scriptsize
        \begin{tabular}{c|c c|c c|c c}
        \toprule
        & \multicolumn{2}{c|}{Cityscapes~(S/8)} & \multicolumn{2}{c|}{COCO-stuff27~(S/8)} & \multicolumn{2}{c}{Potsdam-3~(B/8)} \\
        & HP~\cite{hp} & Ours & HP~\cite{hp} & Ours & HP~\cite{hp} & Ours \\
        \toprule
        Number of positives in $\mathcal{P}$ & 78 & \textbf{1838} & 142 & \textbf{288} & 136 & \textbf{194} \\
        \% of \textit{TP} in $\mathcal{P}$ & 88.83 & \textbf{90.18} & 81.98 & \textbf{87.15} & 83.60 & \textbf{88.62} \\
        \midrule
        Number of negatives in $\mathcal{N}$ & 50K & 43K & 50K & 46K & 50K & 30K \\
        \% of \textit{FP} in $\mathcal{N}$ & 21.33 & \textbf{14.22} & 7.34 & \textbf{5.92} & 34.08 & \textbf{29.01} \\
        \bottomrule
    \end{tabular}
    }
	\label{table_TPFP}
\end{table}
\endgroup

\subsection{Trustworthiness of Positive and Negative Sets.}
If excessive \textit{FP} and \textit{FN} are present in contrastive loss, the model may face undesired attraction and repulsion, respectively. We contend that mitigating these issues can enhance the trustworthiness of contrastive learning.
To verify the robust trustworthiness of our method, previously, we conducted a precision comparison of the positive sets between ours and HP~\cite{hp} on Fig.~\ref{fig_fig1_b}. Notably, ours not only collects more positive samples than HP but also achieves a higher precision~(\textit{i.e.}, ratio of \textit{TP}).
% % To further show the consistency along the datasets,
For further demonstration, we present the ratios of \textit{TP} in the positive set~($\mathcal{P}$) and \textit{FP} in the negative set~($\mathcal{N}$) for three datasets in Tab.~\ref{table_TPFP}.
As shown, ours got a higher ratio of \textit{TP} in $\mathcal{P}$ and lower ratio of \textit{FP} in $\mathcal{N}$ than HP, even with a much larger size of the positive set. For example, we can observe that Ours got 23$\times$ more positives with a higher ratio of \textit{TP}
% for experiments 
on Cityscapes dataset with ViT-S/8 backbone.
Likewise, we ensure the trustworthiness of contrastive learning under more reliable positive and negative sets.

\subsection{Visualization of Patches in Positive Set}

\begin{figure}[t]
    \centering
    {\includegraphics[width=0.96\textwidth]{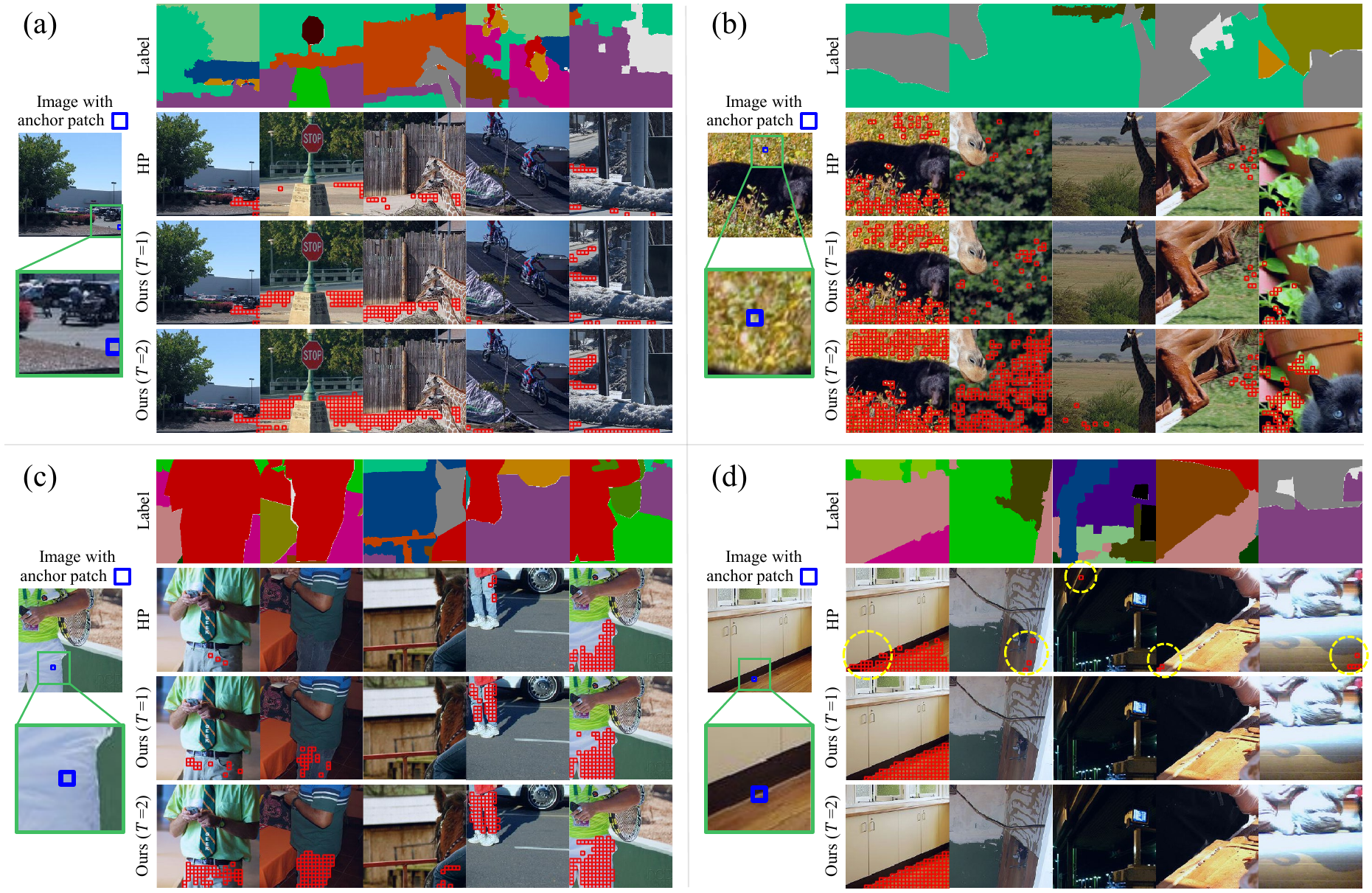} }
    \caption{
    Comparison of gathered positives between HP and PPAP~(Ours) with the visualizations.
    In all examples, blue boxes indicate the selected anchor patch and the red boxes denote the patches that are considered as positive to an anchor.
    In (d), yellow dotted circles exist to highlight the region where the $FP$ are detected.
    }
\label{fig_visualize_positives}
\end{figure}

We illustrate how HP~\cite{hp} and our proposed PPAP organize the positive samples for each anchor through visualizations in Fig.~\ref{fig_visualize_positives}.
As depicted throughout the visualizations, there is a tendency for our method to collect more positive patches compared to the baseline.
On top of that, we can also observe that positive samples that are semantically identical are progressively obtained through the propagation steps in (a), (b), and (c).
Lastly, along with the numerically measured difference in the precision of the positive sets, we find that falsely detected positives by the baseline in (d) are not considered positive in ours.
To account for such a phenomenon, we claim that the presence of ambiguous zones enables the filtering of the hard negative samples. 

\section{Conclusion}
To tackle the challenge of USS, previous approaches have primarily relied on exploiting patch relationships to guide the training process.
In this work, we extend this mainstream to ensure the reliability of the gathered guidance.
Specifically, we consider the data distribution around the anchor to identify densely crowded regions containing samples with similar semantics.
By relocating the proxy anchor to these regions, we expect it to be surrounded by trustworthy positives, creating a large positive set with high precision.
In addition, we address instance-wise ambiguous zones where samples with similar and dissimilar semantics coexist.
By excluding samples from these regions during training, we aim to eliminate \textit{FP} in the negative set, preventing unstable training.
Our state-of-the-art results verify the importance of ensuring the reliability of the supervision in USS.

\noindent\textbf{Acknowledgements.} This work was supported in part by MSIT\&KNPA/KIPoT (Police Lab 2.0, No. 210121M06), MSIT/IITP (No. 2022-0-00680, 2019-0-00421, 2020-0-01821, RS-2024-00437102), and SEMES-SKKU collaboration funded by SEMES.

% \newpage
% \input{6_App1}

\bibliographystyle{splncs04}
\bibliography{main}

\newpage
% \section{Datasets}
% We evaluate our method on COCO-stuff~\cite{coco} and Cityscapes~\cite{cityscapes}, Potsdam-3, and ImageNet-S~\cite{INS} datasets.
% COCO-stuff is a dataset for scene understanding tasks, \textit{e.g.}, semantic segmentation, detection, and image captioning, that consists of 172 classes.
% Among them, the COCO-stuff benchmark for USS utilizes 27 classes.
% Cityscapes is another large-scale dataset for scene understanding that consists of 30 classes captured across 50 different cities.
% Similarly to COCO-stuff, 27 subclasses are used for the benchmark.
% In addition, Potsdam-3 contains satellite images that are divided into 3 classes.
% Lastly, ImageNet-S is a large-scale dataset which has 1.2 million training images with 919 semantic classes.

\section{Limitations.}
% Using unsupervised pretrained features, semantic representations can be misclassified due to minor color differences or edges.
% Furthermore, images taken in close proximity are more prone to mispredictions due to their limited information describing such scenes.
The training guidance derived from patch-wise representation still has limitations in capturing intricate pixel-level details, especially along object edges. This issue becomes more pronounced with larger patch sizes, as ViT-S/16 exhibits lower mIoU compared to ViT-S/8 in this regard.

\section{Inference}
\label{sec_inference}
\begin{figure}[h!]
    {\includegraphics[width=0.5\textwidth]{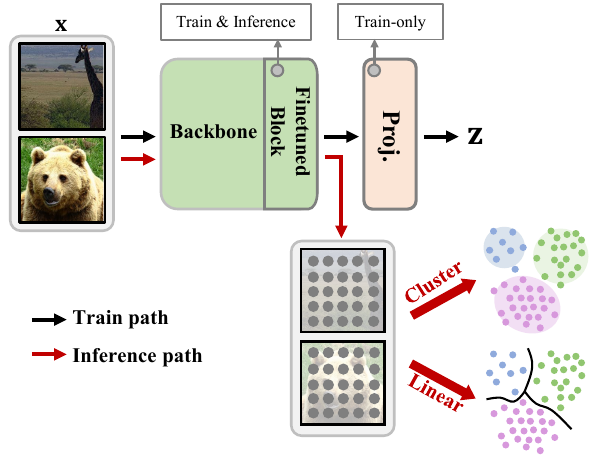} }
    % \vspace{-0.4cm}
    \centering
    \caption{
    Illustration of the training and inference process.
    }
    % \vspace{-0.2cm}
\label{fig_inference_feat}
\end{figure}

We introduce the overall flow of our model at the inference stage in Fig.~\ref{fig_inference_feat}.
During training, we finetune the last block of the Vision Transformer~(ViT) and the projection head which outputs the projected vector $\mathbf{z}$.
While the projected vector $\mathbf{z}$ is used for the training, we use the feature from the backbone for the inference as did in \cite{simclr, swav, dino}.

% At training, we finetune the last block of the Vision Transformer~(ViT) and concurrently train a projection head. 
% This process is guided by a contrastive objective, which is formulated using the projected vector $\mathbf{z}$ based on the criterion obtained from the frozen backbone. During inference, we use the feature extracted directly from the ViT backbone as shown in Fig.~\ref{fig_inference_feat}.

%%%%%%%%%%%%%%%%%%%%%%%%%%%%%%%%%%%%%%%%%%%%%%%%%%%%%%%%%%%%%%%%%%%%%%%%
%%%%%%%%%%%%%%%%%%%%%%%%%%%%%%%%%%%%%%%%%%%%%%%%%%%%%%%%%%%%%%%%%%%%%%%%
\section{Implementation Details}
\label{sec_implementation_details}

In line with existing works~\cite{stego, hp, transfgu}, we utilize a DINO-pretrained ViT as our backbone network. The embedding dimension of the projection layer~($D$) is 384 for ViT-small and 768 for ViT-base.
We set $T$ to 2, 3, and 1 for the experiments on COCO-stuff, Cityscapes, and Potsdam-3 datasets, respectively.
Tab.~\ref{table_hyperparameters} shows the hyperparameter set that yields the best performance for each dataset and backbone.
To reduce the training cost, we use only a random subset of the patch features~\cite{cause}.
Specifically, we use 1/4 of all patch features for the ViT-S/16 backbone and 1/16 for the other backbones.
For the experiments on the ImageNet-S dataset, we use the same hyperparameter settings as for the COCO-stuff dataset.

% to reduce the training cost, a random subset of the patch features from each image are utilized for training. For example, only 1/4 patch features are used for the ViT-S/16 backbone, and 1/16 for the others.

\begingroup
\setlength{\tabcolsep}{6pt} % Default value: 6pt
\renewcommand{\arraystretch}{1.0} % Default value: 1
\begin{table}[h!]
    \centering
    \small
    \scriptsize
    \caption{Hyperparameter settings for each experiment.}
    \begin{tabular}{l| c | c| c | c | c}
        \toprule
        & \multicolumn{2}{c|}{COCO-stuff} &
        \multicolumn{2}{c|}{Cityscapes} & Potsdam-3 \\
        \midrule
        & ViT-S/8 & ViT-S/16 & ViT-S/8 & ViT-B/8 & ViT-B/8 \\
        \midrule
        $\Phi$ & 0.55 & 0.55 & 0.6 & 0.6 & 0.55 \\
        $\Psi$ & 0.2 & 0.15 & 0.2 & 0.2 & 0.15 \\
        $\sigma_\text{pos}$ & 3 & 3 & 3 & 3 & 5 \\
        $\sigma_\text{amb}$ & 3 & 4 & 3 & 2 & 3 \\
        % $T$ & 2 & 2 & 3 & 3 & 1 \\
        % $\tau$ & 0.8 & 0.8 & 0.8 & 0.1 & 0.07 \\
        \bottomrule
    \end{tabular}
    % \vspace{-0.3cm}
    \label{table_hyperparameters}
\end{table}
\endgroup

%%%%%%%%%%%%%%%%%%%%%%%%%%%%%%%%%%%%%%%%%%%%%%%%%%%%%%%%%%%%%%%%%%%%%%%%
%%%%%%%%%%%%%%%%%%%%%%%%%%%%%%%%%%%%%%%%%%%%%%%%%%%%%%%%%%%%%%%%%%%%%%%%
\section{Additional Study}
\label{sec_experimental results}

%%%%%%%%%%%%%%%%%%%%%%%%%%%%%%%%%%%%%%%%%%%%%%%%%%%%%%%%%%%%%%%%%%%%%%%%
\subsection{Study on Effects to Class Frequency}
In dense prediction tasks, the variation in the frequency between different semantics is a very natural phenomenon.
However, this typically leads to the problem of long-tailed data distribution which triggers the large performance gap between classes of high and low frequencies~\cite{longtailseg}.
% There can be variations in the frequency of different semantics in dataset. 
% Due to this long-tailed characteristic of real-world scenario, performance differences can arise based on these frequencies, and generally, semantics with lower frequency tend to have lower performance compared to those with higher frequency~\cite{longtailseg}.
To further analyze the strength of our proposed method, we compare with HP~\cite{hp} on the effects from the perspective of the class frequency.
Specifically, we divide the classes in the COCO-stuff dataset into three groups, i.e., Few, Medium, and Many, according to the data frequency and measure the performances for each group.
Results are reported in Tab.~\ref{table_longtail}.
As shown, we find that our method is particularly notable in learning classes with fewer samples compared to the baseline.
For such a result, we attribute a reason that the number and the precision of the gathered positives are similar between samples as shown in 
Fig.~1 of the main paper.
% Fig.~\ref{fig_fig1}.
% We report the performance based on the class frequency in Tab.~\ref{table_longtail}. This experiment divides the classes of the COCO-stuff dataset into three categories: few, medium, and many, and calculates their average mIoU.
% As shown in Tab.~\ref{table_longtail}, our method shows notable performance improvements in the few classes compared to HP~\cite{hp}. This indicates that the reliability of positive set in HP is low for classes with fewer instances, while our approach demonstrates some success in overcoming this issue.

\begin{table}[h]
\setlength{\tabcolsep}{6.0pt} % Default value: 6pt
\renewcommand{\arraystretch}{1.0} % Default value: 1
\centering
{
    \scriptsize
    \caption{Experimental results considering long-tailed distribution.}
    \begin{tabular}{l | c c c c}
    \toprule
    Method & Few & Medium & Many & All \\
    \midrule
    HP & 28.9 & 47.1 & 75.6 & 42.0 \\ 
    Ours & 32.9 & 50.9 & 76.2 & 45.6 \\
    \bottomrule
    \end{tabular}
    \label{table_longtail}
    }
\end{table}

%%%%%%%%%%%%%%%%%%%%%%%%%%%%%%%%%%%%%%%%%%%%%%%%%%%%%%%%%%%%%%%%%%%%%%%%
\subsection{Different Pretrained Backbones.}
The results in Tab.~\ref{table_different_backbone} demonstrate consistent performance improvements with various backbones pretrained in a self-supervised manner~(\textit{e.g.}, iBoT~\cite{ibot}, SelfPatch~\cite{selfpatch}).
We observed that backbones trained with inter-image relationships consistently enhance performance. However, models such as MAE~\cite{he2022masked} struggle to preserve globally shared semantics in each patch feature across all images~(\textit{e.g.}, 4.3\% of U.mIoU for MAE alone) due to their lack of inter-image relationship modeling~\cite{huang2023contrastive}.

% Main table
\begingroup
\setlength{\tabcolsep}{6pt} % Default value: 6pt
\renewcommand{\arraystretch}{1.0} % Default value: 1
\begin{table}[h]
    \centering
    \caption{Experimental results with various pretrained backbones.${\dagger}$: reproduced.}
    % \vspace{-0.2cm}
    % \small
    \scriptsize
    \begin{tabular}{l | c| c c}
        \toprule
        \multirow{2}{*}{Method}& \multirow{2}{*}{Backbone} & \multicolumn{2}{c}{Unsupervised} \\
        & & Acc. & mIoU \\
        \midrule
        iBoT~\cite{ibot} & ViT-S/16 & 39.2 & 11.8 \\
        ~+~HP$^{\dagger}$~\cite{hp} & ViT-S/16 & 53.2 & 23.0 \\
        \rowcolor{gray!30}
        ~+~PPAP~(Ours) & ViT-S/16 & \textbf{62.4} & \textbf{26.0} \\
        \midrule
        iBoT~\cite{ibot} & ViT-B/16 & 35.7 & 15.0 \\
        ~+~HP$^{\dagger}$~\cite{hp} & ViT-B/16 & 51.1 & 22.4 \\
        \rowcolor{gray!30}
        ~+~PPAP~(Ours) & ViT-B/16 & \textbf{63.4} & \textbf{27.6} \\
        \midrule
        SelfPatch~\cite{selfpatch} & ViT-S/16 & 35.1 & 12.3 \\
        ~+~STEGO~\cite{stego} & ViT-S/16 & 52.4 & 22.2 \\
        ~+~HP~\cite{hp} & ViT-S/16 & 56.1 & 23.2 \\
        \rowcolor{gray!30}
        ~+~PPAP~(Ours) & ViT-S/16 & \textbf{57.8} & \textbf{23.7} \\
        \bottomrule
    \end{tabular}
    \label{table_different_backbone}
\end{table}
\endgroup

%%%%%%%%%%%%%%%%%%%%%%%%%%%%%%%%%%%%%%%%%%%%%%%%%%%%%%%%%%%%%%%%%%%%%%%%
\subsection{Contribution of CRF}
Conditional Random Field~(CRF) utilizes pixel position and RGB color information to smooth the predicted label of each pixel across its neighboring pixels, thereby effectively enhancing the performance of semantic segmentation~\cite{crf}.
Following the previous works~\cite{stego, hp}, we incorporate CRF as a post-processing step in our method. Tab.~\ref{table_crf} presents a performance comparison between HP~\cite{hp} and our method, both with and without the application of CRF.
While the use of CRF leads to performance boosts in all experiments, we highlight the superiority of our PPAP in that it outperforms HP with CRF even without CRF.
% It is evident that in all experiments with both HP and our method, the use of CRF led to performance improvements.
% In particular, a notable observation is that our method, even without applying CRF, outperforms HP which has been enhanced with CRF.
% To show the contribution of the post-processing step, Conditional Random Field~(CRF)~\cite{crf}, we provide the experimental results of our propose method with and without applying CRF compared to HP~\cite{hp} in Tab.~\ref{table_crf}.

% CRF utilizes pixel position and RGB color to smooth the predicted label of each pixel to its neighbors, thereby enhancing semantic segmentation performance. However, there is a risk of smoothing with incorrect labels if the pixels corresponding to an object are generally mispredicted. Such mispredictions occur more frequently in HP, which is why the performance improvement brought by CRF is not as significant as in our method.

% Main table
\begingroup
\setlength{\tabcolsep}{6pt} % Default value: 6pt
\renewcommand{\arraystretch}{1.2} % Default value: 1
\begin{table}[h!]
    \centering
    \small
    \scriptsize
    \caption{Experimental results on COCO-stuff dataset with and without CRF.}
    % \vspace{-0.2cm}
    \begin{tabular}{c | c | c | c c | c c}
        \toprule
        \multirow{2}{*}{Backbone} & \multirow{2}{*}{Method} & \multirow{2}{*}{CRF} & \multicolumn{2}{c|}{Unsupervised} & \multicolumn{2}{c}{Linear} \\
        & & & Acc. & mIoU & Acc. & mIoU \\
        \midrule
        % \rowcolor{gray!30}
        \multirow{4}{*}{ViT-S/8} & \multirow{2}{*}{HP~\cite{hp}} & - & 56.2 & 23.9 & 73.7 & 40.4 \\
        & & \checkmark & 57.2 & 24.6 & 75.6 & 42.7 \\
        \cmidrule{2-7}
        & \multirow{2}{*}{Ours} & - & 57.4 & 26.6 & 76.0 & 45.4 \\
        & & \checkmark & 59.0 & 27.2 & 76.9 & 46.3 \\
        \midrule
        % \rowcolor{gray!30}
        \multirow{4}{*}{ViT-S/16} & \multirow{2}{*}{HP~\cite{hp}} & - & 52.7 & 23.2 & 71.9 & 37.1 \\
        & & \checkmark & 54.5 & 24.3 & 74.1 & 39.1 \\
        \cmidrule{2-7}
        & \multirow{2}{*}{Ours} & - & 59.9 & 25.0 & 74.3 & 42.1 \\
        & & \checkmark & 62.9 & 26.5 & 76.0 & 43.3 \\
        \bottomrule
    \end{tabular}
    \label{table_crf}
    % \vspace{-0.5cm}
    % \vspace{-0.1cm}
    % \vspace{-0.4cm}
\end{table}
\endgroup

%%%%%%%%%%%%%%%%%%%%%%%%%%%%%%%%%%%%%%%%%%%%%%%%%%%%%%%%%%%%%%%%%%%%%%%%
\subsection{Qualitative Results without CRF}
\begin{figure*}
    \centering
    {\includegraphics[width=0.85\textwidth]{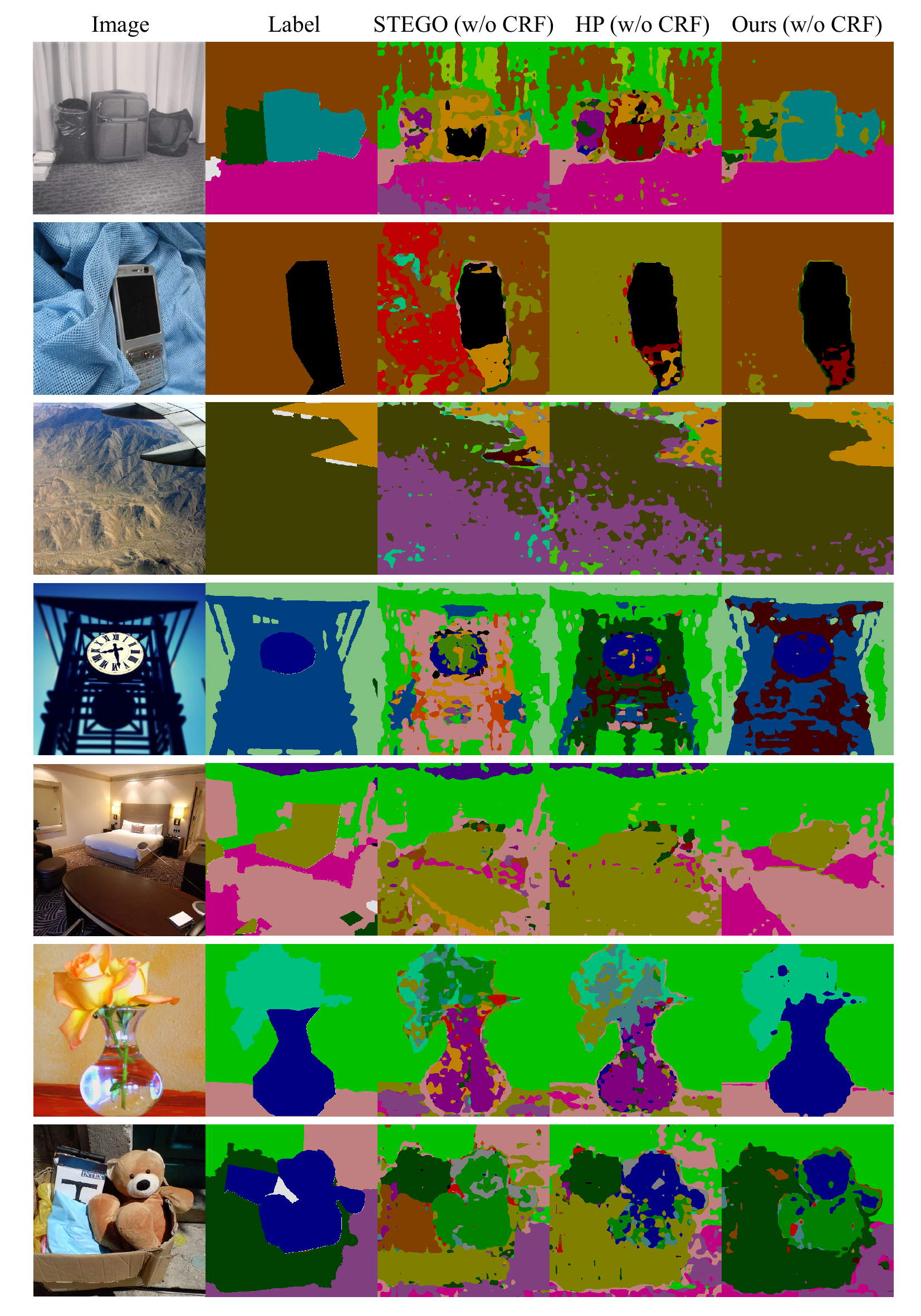} }
    % \vspace{-0.1cm}
    \caption{
    Qualitative results without CRF.
    }
    % \vspace{-0.5cm}
\label{fig_model_qual_nocrf1}
\end{figure*}

\begin{figure*}
    \centering
    {\includegraphics[width=0.85\textwidth]{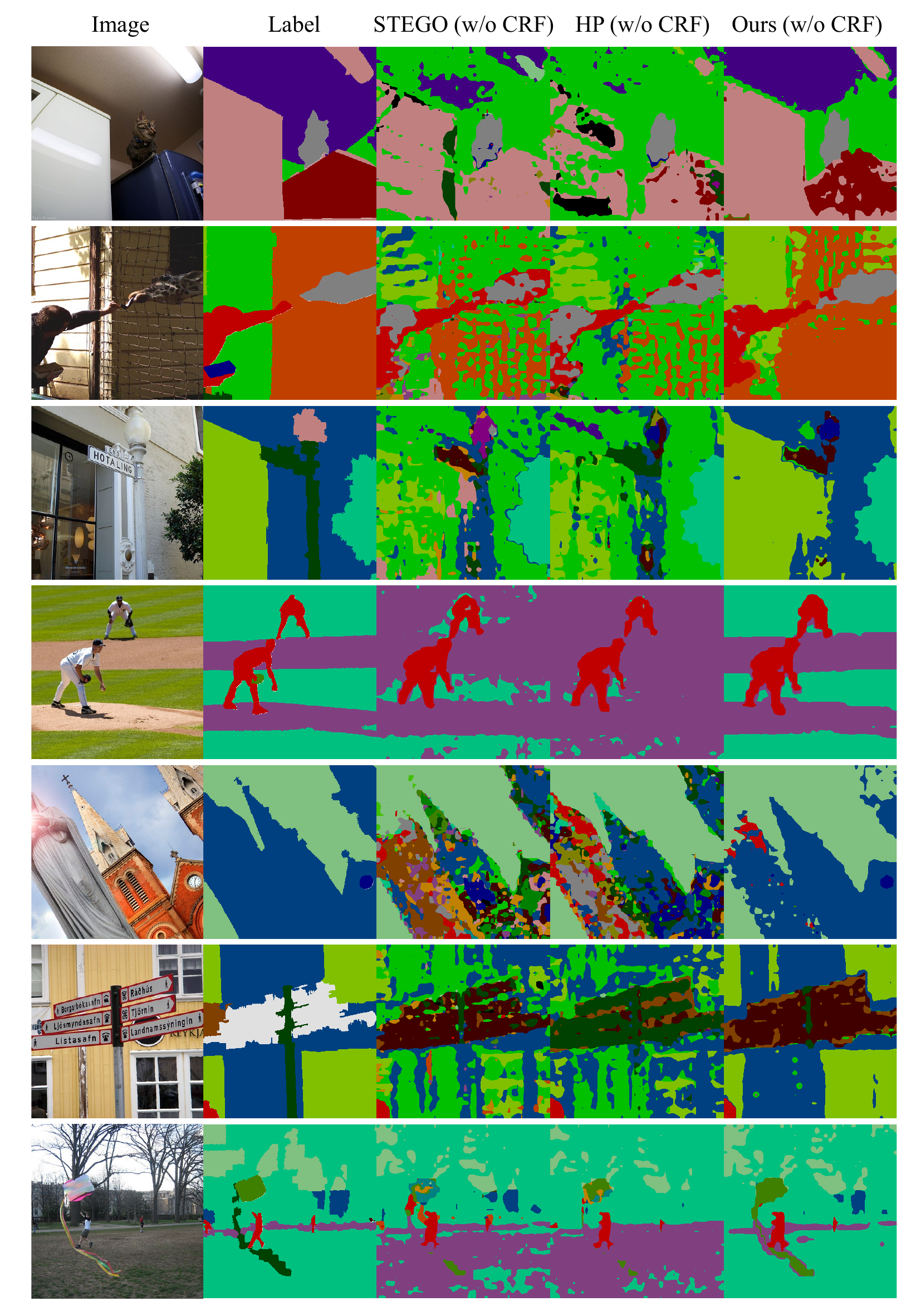} }
    % \vspace{-0.1cm}
    \caption{
    Qualitative results without CRF.
    }
    % \vspace{-0.5cm}
\label{fig_model_qual_nocrf2}
\end{figure*}

We provide visualizations comparing the predictions of our proposed PPAP with those of existing methods, \textit{i.e.}, STEGO~\cite{stego}, and HP~\cite{hp}, without CRF.
% For the model with CRF, the visualization results are presented in Fig.~\ref{fig_model_qual_crf1}, \ref{fig_model_qual_crf2} and \ref{fig_model_qual_crf3}. 
The results without CRF are shown in Fig.~\ref{fig_model_qual_nocrf1} and \ref{fig_model_qual_nocrf2}.
As can be noticed, we point out that existing methods tend to be prone to noise, whereas our method demonstrates robustness against pixel-wise noises even without applying the CRF.

% %%%%%%%%%%%%%%%%%%%%%%%%%%%%%%%%%%%%%%%%%%%%%%%%%%%%%%%%%%%%%%%%%%%%%%%%
% %%%%%%%%%%%%%%%%%%%%%%%%%%%%%%%%%%%%%%%%%%%%%%%%%%%%%%%%%%%%%%%%%%%%%%%%
% \section{Limitations.}
% % Using unsupervised pretrained features, semantic representations can be misclassified due to minor color differences or edges.
% % Furthermore, images taken in close proximity are more prone to mispredictions due to their limited information describing such scenes.
% The training guidance derived from patch-wise representation still has limitations in capturing intricate pixel-level details, especially along object edges. This issue becomes more pronounced with larger patch sizes, as ViT-S/16 exhibits lower mIoU compared to ViT-S/8 in this regard.

\end{document}